\documentclass{article}

\usepackage{svg}
\usepackage{microtype}
\usepackage{graphicx}
\usepackage{subcaption}
\usepackage{booktabs} 
\usepackage{caption}
\usepackage{mwe}
\usepackage[table]{xcolor}  
\usepackage{tabularx}
\usepackage{enumitem}
\usepackage{multirow}
\usepackage{rotating}

\usepackage[utf8]{inputenc}
\usepackage[table]{xcolor}
\usepackage{longtable}
\usepackage{array}

\usepackage{hyperref}

\usepackage[most]{tcolorbox} 
\definecolor{mybg}{HTML}{FAE5D8}    
\definecolor{myframe}{HTML}{180018} 

\tcbset{
    myfancybox/.style={
        colback=mybg,       
        colframe=myframe,      
        coltitle=white,        
        fonttitle=\bfseries\normalsize, 
        fontupper=\small,      
        boxrule=0pt,           
        toprule=3pt,           
        bottomrule=3pt,        
        leftrule=0pt,
        rightrule=0pt,
        arc=4pt,               
        left=6pt,
        right=6pt,
        top=4pt,
        bottom=4pt,
        title={Your Box Title Here} 
    }
}

\usepackage[preprint]{icml2026}

\usepackage{amsmath}
\usepackage{amssymb}
\usepackage{mathtools}
\usepackage{amsthm}

\usepackage[capitalize,noabbrev]{cleveref}

\theoremstyle{plain}

\theoremstyle{definition}

\theoremstyle{remark}

\usepackage[textsize=tiny]{todonotes}

\usepackage{acronym}
\newacro{LLMs}{Large Language Models}
\newacro{MHA}{Multi-Head Attention}
\newacro{FFN}{Feed Forward Network}
\newacro{GQA}{Grouped-Query Attention}

\newcommand{\Dart}{\textsc{Dart}\xspace}

\usepackage{multirow}
\usepackage{makecell}

\icmltitlerunning{\Dart-ing Through the Drift: Dynamic Tracing of Knowledge Neurons for Adaptive Inference-Time Pruning}

\begin{document}

\twocolumn[
  \icmltitle{\Dart-ing Through the Drift: Dynamic Tracing of Knowledge Neurons for Adaptive Inference-Time Pruning}



  \icmlsetsymbol{equal}{*}

  \begin{icmlauthorlist}
    \icmlauthor{Abhishek Tyagi}{equal,nus}
    \icmlauthor{Yunuo Cen}{equal,nus}
    \icmlauthor{Shrey Dhorajiya}{bits}
    \icmlauthor{Bharadwaj Veeravalli}{nus}
    \icmlauthor{Xuanyao Fong}{nus}
  \end{icmlauthorlist}

  \icmlaffiliation{nus}{Department of Electrical and Computer Engineering, National University of Singapore, Singapore}
  \icmlaffiliation{bits}{Department of Computer Science, Birla Institute of Technology and Science, Pilani, India}
  \icmlcorrespondingauthor{Xuanyao Fong}{kelvin.xy.fong@nus.edu.sg}

  \icmlkeywords{Machine Learning, ICML}

  \vskip 0.3in
]



\printAffiliationsAndNotice{\textsuperscript{*}Equal contribution. These authors are listed in alphabetical order. The order should not be used to determine the extent of authors' contributions.}

\begin{abstract}
  Large Language Models (LLMs) exhibit substantial parameter redundancy, particularly in Feed-Forward Networks (FFNs).
  Existing pruning methods suffer from two primary limitations.
  First, reliance on dataset-specific calibration introduces significant data dependency and computational overhead.
  Second, being predominantly static, they fail to account for the evolving subset of knowledge neurons in LLMs during autoregressive generation as the context evolves.
  To address this, we introduce \Dart, \emph{i.e.}, \emph{Dynamic Attention-Guided Runtime Tracing}), a lightweight, training-free method that performs on-the-fly context-based pruning. 
  \Dart monitors shifts in attention score distributions to infer context changes, dynamically updating neuron-level masks to retain salient parameters.
  Across ten benchmarks, \Dart outperforms prior dynamic baseline, achieving accuracy gains of up to 14.5\% on \textsc{LLaMA-3.1-8B} at 70\% FFN sparsity.
  Furthermore, \Dart achieves up to 3$\times$ better ROUGE-L scores with respect to static-masked pruning on summarization tasks, with its performance comparable to the original dense models. 
  We conclusively demonstrate that the proposed framework effectively adapts to diverse semantic contexts, preserves model capabilities across both general and domain-specific tasks while running at less than 10MBs of memory for \textsc{Llama-3.1-8B}(16GBs) with 0.1\% FLOPs overhead.
  The code is available at \url{https://github.com/seeder-research/DART}.
\end{abstract}

\section{Introduction}
\ac{LLMs} are fundamental in modern artificial intelligence and demonstrate strong capabilities across diverse applications, including code generation~\cite{chen2021evaluating}, mathematical reasoning~\cite{wei2022chain}, biomedical diagnostics~\cite{singhal2023large}, and embodied robotics~\cite{driess2023palme}.  
However, this capability is largely driven by the scaling law~\cite{kaplan2020scaling,hoffmann2022an}, resulting in models with trillions of parameters that incur prohibitive memory and computation overhead~\cite{chowdhery2023palm}. 
Consequently, deploying state-of-the-art models poses a fundamental challenge in achieving high throughput and low latency simultaneously, particularly under the diverse and dynamic workloads encountered in real-world inference~\cite{yu2022orca}.
This challenge has motivated a range of techniques to improve inference efficiency without sacrificing model quality~\cite{wan2023efficient}.
At the hardware level, approaches such as kernel optimization~\cite{dao2022flashattention} and efficient batching~\cite{kwon2023efficient} seek to utilize hardware resources better and amortize computation across requests.
At the algorithm level, architectural modifications and compression techniques, including quantization~\cite{dettmers2022gpt3} and pruning~\cite{frantar2023sparsegpt}, aim to reduce the computational overhead of inference.



Most existing pruning methods rely on dataset- or model-specific calibration and impose static sparsity patterns.
Early approaches such as \textsc{Wanda}~\cite{sun2024a} and \textsc{SparseGPT}~\cite{frantar2023sparsegpt} identify parameters that are redundant on average over a calibration set and apply a single fixed mask across all inputs. Subsequent methods, including DLP~\cite{chen2025dlp} and OWL~\cite{10.5555/3692070.3694428}, follow the same paradigm by computing global or layer-wise importance scores from calibration data and permanently pruning parameters based on these scores.
While static pruning reduces inference cost, it fails to capture the input-dependent computation of \ac{LLMs}, where parameters that are rarely used during calibration can be critical for specific prompts at inference.

Results in \citet{geva2021transformer}, \citet{dai2022knowledge}, and \citet{meng2022locating} show that parameter importance varies with the input knowledge domain, revealing strong contextual sparsity in which only a small subset of parameters is active per token and motivating the development of dynamic sparsification methods.
\textsc{DejaVu}~\cite{liu2023deja} predicts token-level importance to skip inactive attention heads or \ac{FFN} neurons, but requires auxiliary models trained for specific datasets and architectures.
\textsc{Griffin}~\cite{dong2024prompt} avoids additional training by reusing highly activated \ac{FFN} neurons from the prefill stage, but mainly exploits intra-layer sparsity and does not account for inter-layer interactions that shape knowledge representation.
These limitations suggest that effective sparsification should be both layer-specific and context-aware.

\begin{figure}
    \centering
    \includegraphics[width=0.8\linewidth]{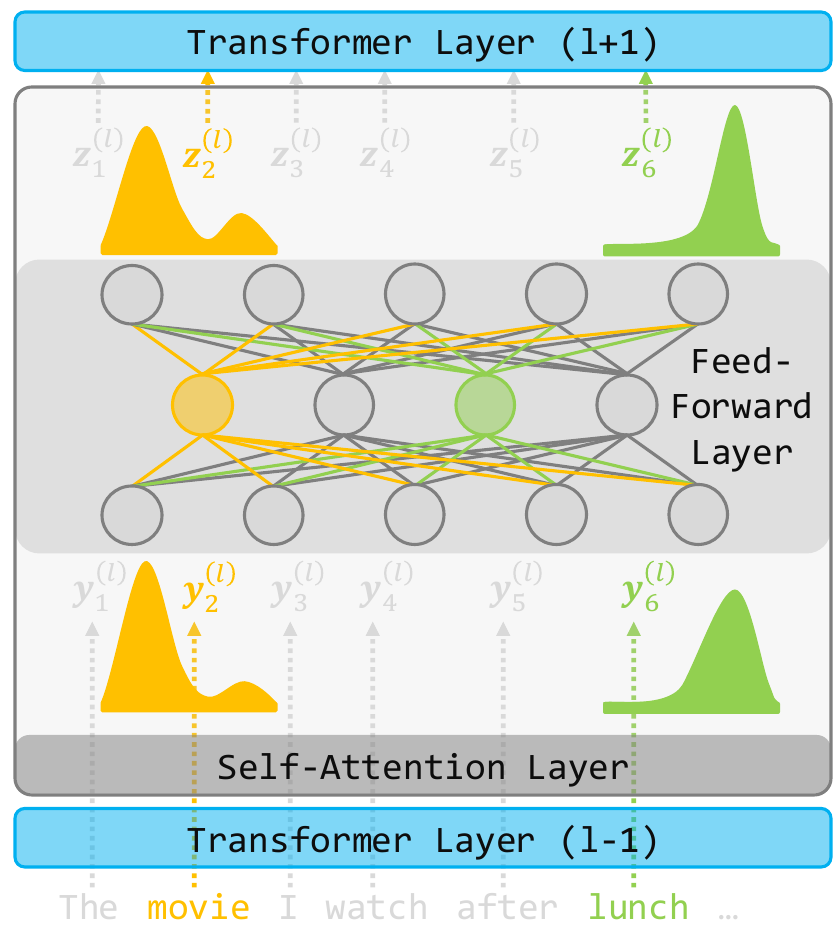}
    \caption{Variation in attention output distributions across different input contexts. 
    This distributional shift necessitates the activation of distinct neuron subsets.}
    \label{fig:knowledge_neuron}
\end{figure}

Furthermore, we identify a critical failure mode in prompt-calibrated pruning, which we term as \emph{knowledge drift}.
As the generation progresses, the model’s internal representations evolve in response to newly generated tokens, causing the relevance of specific neurons to shift over time.
As illustrated in \cref{fig:knowledge_neuron}, neurons being inactive during prefill may later become critical for expressing factual details of domain-specific knowledge. 
Permanently pruning such neurons can therefore bias the generation process, where certain facts or concepts become increasingly underrepresented over long-horizon generation.

To address these challenges, we introduce \Dart, \emph{i.e.}, \emph{Dynamic Attention-Guided Runtime Tracing}, a lightweight pruning framework for \ac{LLMs} that effectively identifies the knowledge housed within the different neurons in each layer and also adjusts the set of selected knowledge neurons at runtime, thus mitigating bias introduced by pruning. 
Our framework comprises two primary components.
First, we propose a context-aware neuron selector that allocates sparsity budgets based on the intrinsic knowledge density of each \ac{FFN} layer. 
By modeling relative intra-layer importance and inter-layer interactions, this module performs structured pruning without auxiliary training, enhancing inference throughput while preserving \ac{LLMs}' performance.
However, during long-horizon generation, masks derived from a static prefix often fail as the semantic context evolves. 
To address this issue, we further introduce a context switch detector that monitors the distributional shift of attention layer outputs relative to the reference context.
When the divergence between current activations and the reference centroids exceeds a threshold, the framework triggers a mask update to reinstate critical parameters.
Empirical evaluations on multi-topic generation and summarization tasks demonstrate that this mechanism effectively recovers the predictive capabilities typically lost to static pruning methods.
Specifically, across ten benchmarks, \Dart outperforms prior dynamic baselines, achieving up to +14.5 accuracy gains on \textsc{LLaMA-3.2-3B} and up to +19.6 on \textsc{LLaMA-3.1-8B} at 70\% FFN sparsity. 
Furthermore, \Dart is shown to achieve up to 3$\times$ better ROUGE-L scores with respect to static-masked pruning on summarization tasks, with its performance comparable to the original dense models.
Overall, the contributions are summarized as follows.

\textbf{A lightweight, model-agnostic pruning method} that derives per-layer masks for the \ac{FFN} sublayers. 
By modeling the geometric contribution of each layer to the residual stream, this method achieves deterministic inference speedups while preserving accuracy.

\textbf{A failure mode termed as \emph{knowledge drift} is identified} in dynamic pruning. 
In this mode, static masks derived from an initial input prefix fail to support the evolving semantic requirements of autoregressive generation. 
This leads to error accumulation during long-horizon tasks.

\textbf{An online \emph{knowledge drift} detector} that monitors distributional shifts in attention outputs.
This detector acts as a runtime supervisor, triggering adaptive mask updates to resolve semantic misalignment.

\textbf{Verification through extensive benchmarks} across zero-shot and multi-shot domain-specific and natural language processing datasets, as well as multi-topic summarization. 
Empirical results demonstrate that \Dart significantly reduces the performance gap between pruned and dense models. It outperforms existing static, dynamic, structured, and unstructured pruning baselines.

\section{Preliminary}
\subsection{Autoregressive Generation and Transformer}
We consider a transformer-based LLM with $L$ layers.
Given an input sequence of tokens, the model generates tokens autoregressively.
At each time step $t$, each layer $l$ takes the hidden states $\mathbf{X}^{(l)} \in \mathbb{R}^{t \times d}$ from the previous layer as input, where $d$ is the hidden-state dimension.
Each transformer layer consists of two primary sub-modules: \ac{MHA} and \ac{FFN}, along with residual connections.

\paragraph{Multi-Head Attention.} 
For a specific attention head $h$, the current-token embedding $x^{(l)}_t\in\mathbb{R}^{1 \times d}$ is projected to a query using $\mathbf{W}_Q^{(h)}\in \mathbb{R}^{d \times d_h}$.
The hidden states $\mathbf{X}^{(l)}$ (current and previous tokens) are projected to keys and values using $\mathbf{W}_K^{(g)}, \mathbf{W}_V^{(g)}\in \mathbb{R}^{d \times d_h}$.
In modern \ac{LLMs}, \ac{GQA} is often employed, where multiple query heads $h\in f(g)$ share the same key and value matrices within each group $g$~\cite{ainslie2023gqa,touvron2023llama}.
The attention head is computed as:
\begin{equation*}
    \mathbf{A}^{(h)} = \text{Softmax}\left(\frac{x^{(l)}_t\mathbf{W}_Q^{(h)}  (\mathbf{X}^{(l)}\mathbf{W}_K^{(g)} )^\top}{\sqrt{d_h}}\right) \mathbf{X}^{(l)}\mathbf{W}_V^{(g)}.
\end{equation*}
Eventually, the \ac{MHA} output is the projection of the concatenated attention head via $\mathbf{W}_O^{(l)}\in\mathbb{R}^{n_hd_h \times d}$.
\begin{equation*}
    \mathbf{O}^{(l)} = \text{Concat}\left(\mathbf{A}^{(1)},\ldots,\mathbf{A}^{(n_h)}\right)\mathbf{W}_O^{(l)}
\end{equation*}

\paragraph{Feed Forward Network.}
The \ac{FFN} operates on each token independently.
Given $\mathbf{y}^{(l)}_t\in\mathbb{R}^d$ as input, modern \ac{LLMs} commonly use gated \ac{FFN}s~\cite{shazeer2020glu} with an up-projection matrix and a gate-projection matrix, $\mathbf{W}^{(l)}_{\mathrm{up}},\mathbf{W}^{(l)}_{\mathrm{gate}}\in\mathbb{R}^{d\times m}$:
\begin{equation}
\mathbf{u}^{(l)}_t = \mathbf{y}^{(l)}_t \mathbf{W}^{(l)}_{\mathrm{up}} \odot \mathrm{SiLU}\left(\mathbf{y}^{(l)}_t \mathbf{W}^{(l)}_{\mathrm{gate}}\right),\label{eq:ffn-up}
\end{equation}
followed by a down-projection matrix $\mathbf{W}^{(l)}_{\mathrm{down}}\in\mathbb{R}^{m\times d}$:
\begin{equation}
\mathbf{z}^{(l)}_t = \mathbf{u}^{(l)}_t \mathbf{W}^{(l)}_{\mathrm{down}}.\label{eq:ffn-down}
\end{equation}
With the residual connection, the \ac{FFN} block output can be written as
\begin{equation}
    \mathbf{z}^{(l)}_t = \mathbf{y}^{(l)}_t + \text{FFN}^{(l)}(\mathbf{y}^{(l)}_t), \label{eq:ffn}
\end{equation}
where $\text{FFN}^{(l)}$ denotes the $l$-th layer \ac{FFN}.

\subsection{Contextual Sparsity}
Modern \ac{LLMs} adopt architectural choices that significantly reshape the parameter distribution across submodules. 
With \ac{GQA}, multiple query heads share key and value projections, reducing the parameters of the attention mechanism relative to standard \ac{MHA}. 
In contrast, the use of the gated \ac{FFN}, with an additional gate projection, substantially increases the computational cost.
As a result, efficiency gains from pruning are increasingly governed by decisions made in the \ac{FFN}, rather than in attention layers alone.

\paragraph{Embedding clustering in MHA is less contextually sparse.}
It was hypothesized~\cite{liu2023deja} that the attention head can be regarded as a mean-shift step to push input embeddings of different tokens together if they are already neighbors in a projection space specified by $\mathbf{W}_Q^{(h)}(\mathbf{W}_K^{(g)})^\top$. 
Different heads learn different projection spaces to perform clustering. 
Given a set of token embeddings within a certain context tends to have high attention scores in certain attention heads, \emph{i.e.}, the rest of the heads are hardly activated. 
However, in \ac{GQA}, multiple query heads share the same key projection, which enforces them to operate within a shared subspace.
Consequently, the intrinsic sparsity of the attention mechanism is compromised, yielding a denser pattern where fewer heads remain inactive.

\paragraph{Knowledge neurons in FFN are more contextually sparse.}
An \ac{FFN} in \ac{LLMs} is regarded as a knowledge lookup table with $m$ entries, and each entry encodes factual knowledge~\textcolor{blue}~\cite{geva2021transformer}. 
\cref{eq:ffn-down} can be interpreted as, every neuron in $\mathbf{u}_t^{(l)}\in\mathbb{R}^m$ retrieve the knowledge in the knowledge database $\mathbf{W}^{(l)}_{\mathrm{down}}$, and $\mathbf{z}_t^{(l)}\in\mathbb{R}^d$ encodes the knowledge as the linear combination of all entries. 
Given a set of token embeddings within a knowledge domain, only the neurons corresponding to that domain will be activated, \emph{i.e.}, have higher activation values~\cite{meng2022locating}. 

\section{Theoretical Framework}
In this section, we present our theoretical framework, termed \Dart.
We begin by formalizing a lightweight, context-based dynamic pruning mechanism for \ac{FFN}s
Recognizing that semantic redundancy is not uniform across the network, we effectively demonstrate that an optimal pruning ratio varies by layer and propose a sensitivity-aware allocation strategy to determine layer-wise budgets
However, we observe that autoregressive generation is not static--it frequently transitions across distinct knowledge domains, \emph{i.e.}, a different set of neurons should be activated.
We show that this \emph{knowledge drift} is strictly reflected in the outputs of preceding attention layers.
Finally, leveraging this insight, we introduce the \emph{Knowledge Tracer}: a mechanism that monitors attention dynamics to detect domain shifts and invoke the pruner only when necessary.

\subsection{Recap of Context-Based Pruning}
In the \Dart framework, we hypothesize that for any given token generation step, only a subset of neurons is required to maintain semantic fidelity.
A similar observation was previously made in \cite{dong2024prompt}. 
We introduce a context-dependent binary mask $M^{(l)} \in \{0, 1\}^m$.
The pruned \ac{FFN} operation is, thus, defined as an alternative to Eq.~\eqref{eq:ffn-down}, by:
\begin{equation*}
    \hat{\mathbf{z}}_{t}^{(l)} = (M^{(l)} \odot u_t^{(l)})\mathbf{W}_{down}^{(l)},
\end{equation*}
where $\odot$ denotes the Hadamard product.
To capture the temporal significance of each neuron, we do not rely on instantaneous activation at a single step. 
We compute a cumulative importance score, $s_i$, for the $i$-th neuron by aggregating activation magnitudes over a window of $\tau$ tokens.
\begin{align*}
    s_i = \sum_{t=1}^\tau \left(u^{(l)}_{t,i}\right)^2
\end{align*}
We then generate a binary mask $M^{(l)}$ that selects the top-$k$ neurons with the highest cumulative scores.
Formally, 
\begin{align*}
    M_{i}^{(l)}= \mathbb{I}\left[\left|\left\{s_j \mid s_j>s_i\right\}\right|<k\right],
\end{align*}
where $\mathbb{I}$ is the indicator function.
Once computed at step $\tau$, the mask effectively prunes the layer by zeroing out the $w_{gate,(j,i)}^{(l)}\in W_{gate}^{(l)}$, $w_{up,(j,i)}^{(l)}\in W_{up}^{(l)}$, and $w_{down,(i,j)}^{(l)}\in W_{down}^{(l)}$ for all $M_{i}^{(l)}=0$.

\subsection{Layer-Wise Contextual Sparsity}
\begin{figure}[t]
  \begin{center}
    \centerline{\includegraphics[width=\columnwidth]{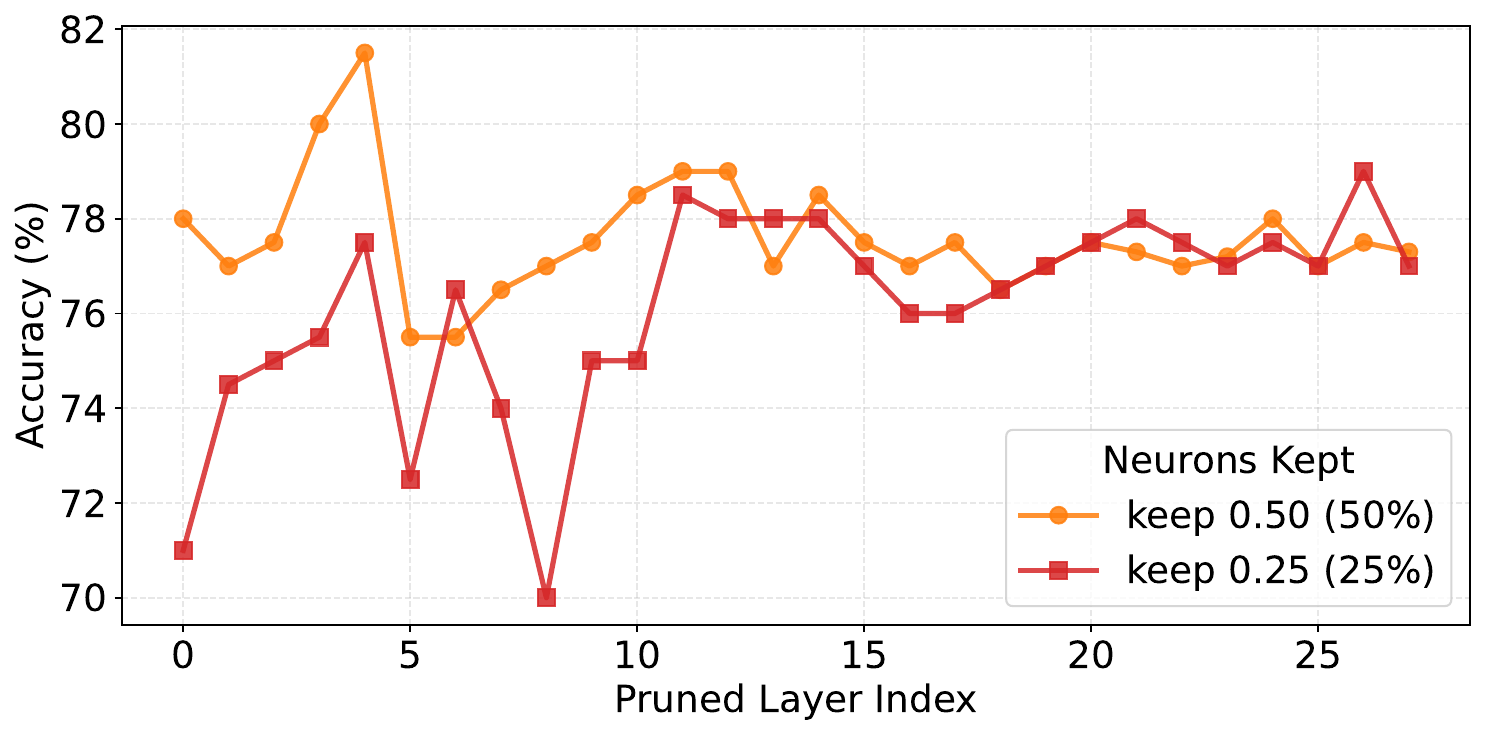}}
    \caption{Layerwise sensitivity of \textsc{LLaMA-3.2-3B} on the MMLU marketing subject.}
    \label{fig:single_layer_prune}
  \end{center}
\end{figure}

Having established the theoretical basis for \emph{intra-layer} neuron pruning, we now address the determination of \emph{inter-layer} sparsity ratios.
The \ac{FFN}s of \ac{LLMs} are characterized in \cite{geva2021transformer} as repositories of factual knowledge, where individual neurons act as knowledge entries that are selectively activated depending on the input context. 
It indicates the number of neurons that can be pruned is inherently heterogeneous across layers, which is implied in \cref{fig:single_layer_prune}.

\paragraph{Layer Sensitivity.}
To estimate layer importance efficiently at runtime, we measure the change each layer induces on the token embeddings.
We hypothesize that a layer is functionally important if its output significantly alters the direction or magnitude of the input state.
We proposed a sensitivity score $S^{(l)}_t$ to quantify the impact of $\text{FFN}^{(l)}$, \emph{i.e.}, \cref{eq:ffn}, with respect to the $t$-th token by
\begin{equation}\label{eq:sensitivity}
S^{(l)}_t = \left(1 - \frac{\mathbf{y}^{(l)}_{t}\mathbf{z}^{(l)}_{t}}{\|\mathbf{y}^{(l)}_{t}\|\cdot\|\mathbf{z}^{(l)}_{t}\|}\right) \cdot \frac{\|\mathbf{z}^{(l)}_{t} - \mathbf{y}^{(l)}_{t}\|}{\|\mathbf{y}^{(l)}_{t}\|},
\end{equation}
where the first term measures the change in direction (via cosine similarity), while the second term measures the intensity of the update relative to the input.
Hence, layers exhibiting substantial directional deviation or high-intensity updates are assigned higher sensitivity scores, indicating their semantic criticality.
For a sequence of input tokens of length $\tau$, the average sensitive score is
\begin{equation*}
    \bar{S}^{(l)}_{t,\tau} = \frac{1}{\tau}\sum_{i=0}^{\tau-1}{S}^{(l)}_{t+i}.
\end{equation*}
Inspired by~\cite{chen2025dlp}, we normalize these scores across all $L$ layers to produce a relative importance 
\begin{equation*}
I^{(l)}_{t,\tau} = 1-\frac{\bar{S}^{(l)}_{t,\tau}}{\sum_{j=0}^{L-1} \bar{S}^{(j)}_{t,\tau}}. 
\end{equation*}

\begin{figure}[t]
  \begin{center}
    \centerline{\includegraphics[width=\columnwidth]{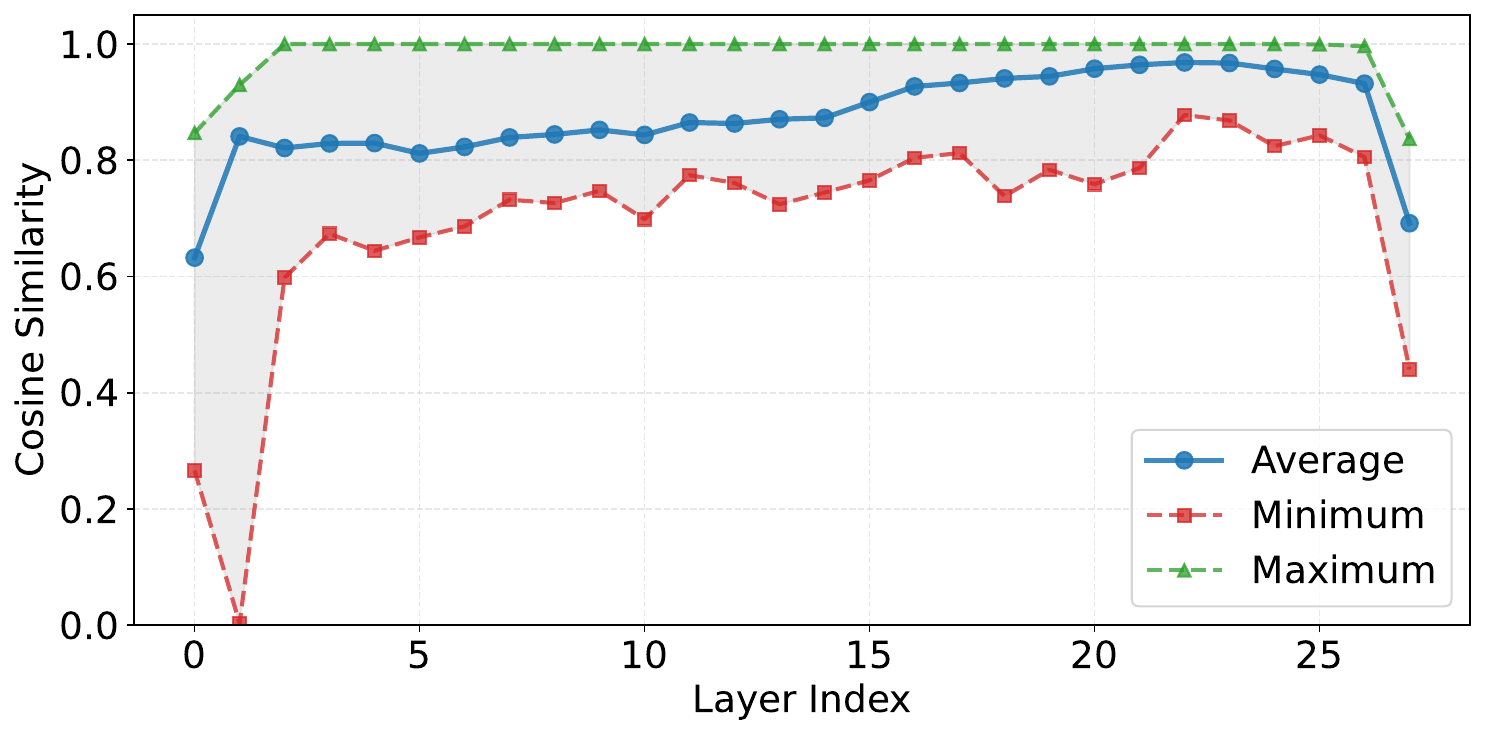}}
    \caption{Cosine similarity of embeddings pre- and post-FFN layers in \textsc{LLaMA-3.2-3B} for the MMLU marketing subject.}
    \label{fig:cosine_sim}
  \end{center}
\end{figure}

\paragraph{Depth-Aware Sparsification.}
Observations on layer-wise redundancy indicate that the first and last layers are functionally denser and less robust to pruning compared to the deep intermediate blocks~\citep{gromov2024unreasonable, men2025shortgpt}.
As shown in \cref{fig:cosine_sim}, earlier layers consistently induce larger embedding changes than later layers. 
This aligns with the pruning results in \cref{fig:single_layer_prune}, where pruning later layers leads to substantially lower accuracy degradation compared to pruning earlier layers. 
Let $\tilde{l} = l / (L-1)$ be the normalized layer depth. 
For each layer $l\in[0, L-1]$, we employ a depth-dependent factor $D^{(l)}$ as
\begin{equation}\label{eq:depth}
    D^{(l)} = \min\left(1, \alpha_e + (1-\alpha_e)\frac{\tilde{l}}{\beta_e}, \alpha_l + (1-\alpha_l)\frac{1-\tilde{l}}{\beta_l}\right),
\end{equation}
where $\alpha_e, \alpha_l \in (0, 1)$ are the scaling factors, and $\beta_e, \beta_l$ represent the normalized widths of the early and late intervals, subject to $\beta_e + \beta_l \leq 1$.


\paragraph{Iterative Sparsity Redistribution.}
With $ I^{(l)}_{t,\tau}$ and $D^{(l)}$, the pruning ratio $p^{(l)}$ for each layer is solved as a constrained allocation problem. 
Given a global target sparsity $\rho\in[0,1]$, the total pruning budget is $\Delta_0 = \rho \cdot L$. 
To ensure $p^{(l)}\in\left[p_{\min}, p_{\max}\right]$, we employ an iterative redistribution algorithm.
We initialize the pruning ratios $p^{(l)}_0 = 0$ and the active set $A_0 = \{0, \dots, L-1\}$.
At each iteration $k$, we distribute the remaining budget $\Delta_{k-1}$ among the active layers in $\mathcal{S}_{t-1}$:
\begin{align}
\nonumber p^{(l)}_k &= \text{clip}\left(p^{(l)}_{k-1} + \frac{\Delta_{k-1} \cdot I^{(l)}_{t,\tau} \cdot D^{(l)}}{\sum_{j\in A_{k-1}} I^{(j)}_{t,\tau} \cdot D^{(j)}},p_{\min},p_{\max}\right),\\
\label{eq:iter_redist}\Delta_{k} &= \Delta_{k-1} - \sum_{j\in A_{k-1}} \left(p_k^{(l)}-p_{k-1}^{(l)}\right), \\
\nonumber A_k &= \left\{j|p_k^{(j)}\in(p_{\min}, p_{\max})\right\}.
\end{align}
This process repeats when $\Delta_{k}=0$ and the final pruning ratio $p^{(l)} = p^{(l)}_k$. 

\begin{figure*}
    \centering
    \includegraphics[width=0.95\linewidth]{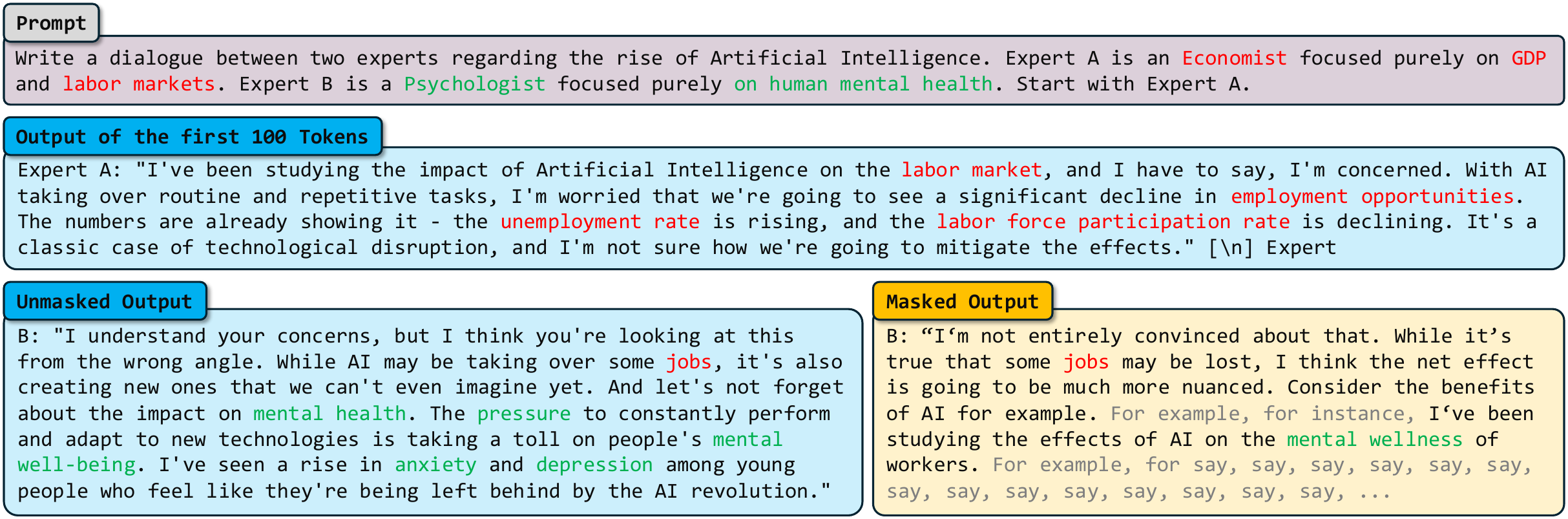}
    \caption{Comparison of unmasked and masked model outputs for a multi-expert dialogue. The pruning mask generated for a specific topic biases the output of another topic.}
    \label{drift_text}
\end{figure*}

\begin{figure}[t]
\begin{center}
\includegraphics[width=\columnwidth]{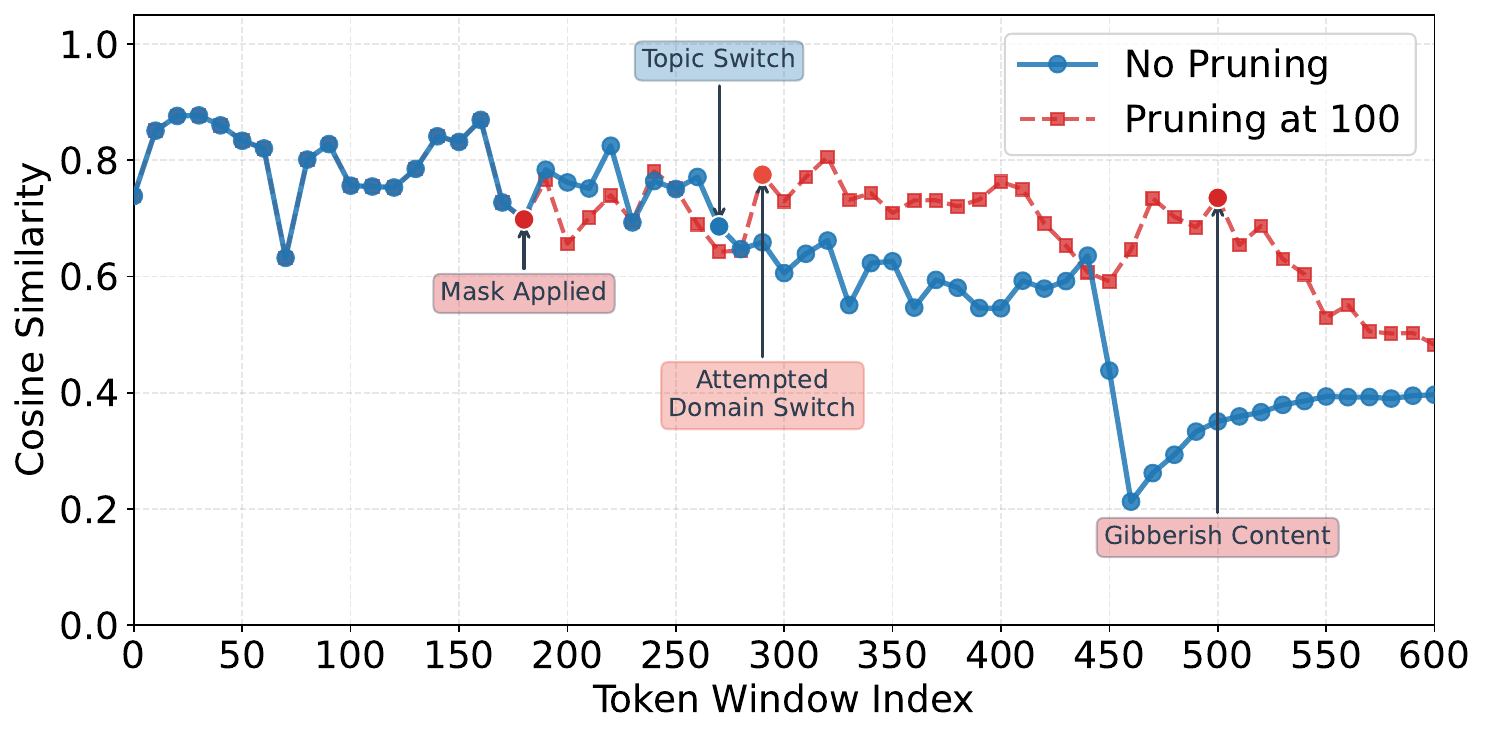}
\caption{Cosine similarity of the attention vector over time for unpruned and pruned \ac{FFN} at the last layer.}
\label{fig:unmasked_masked}
\end{center}
\end{figure}

\subsection{Knowledge Drift}
The pruning strategy described above derives sparsity masks conditioned solely on the input prefix.
However, LLM generation is an autoregressive process where the semantic context evolves with each newly generated token.
Consequently, the specific subset of neurons required to represent the hidden state is not static.
As the LLM generation progresses, the model may require knowledge neurons that were pruned.
As shown in \cref{drift_text}, when such a transition occurs, the masked model produces incoherent output, while the unmasked model remains coherent.
This indicates that fixed prompt-derived masks do not generalize across knowledge-domain shifts.



\paragraph{Observation of Knowledge Drift.}
To quantify this phenomenon, we analyze the trajectory of attention output vectors from the final layer.
We compute the cosine similarity between the mean attention vector within a sliding window of ten tokens and the reference mean vector derived from the tokens used for mask construction.
\cref{fig:unmasked_masked} illustrates the resulting similarity trajectories. 
In the unpruned baseline, we observe a sharp decline between tokens 425 and 475, indicating a transition to a distinct semantic context.
In the pruned model, the cosine similarity remains above 0.6.
This suggests that although the attention mechanism attempts to steer the generation toward the new context, the pruned \ac{FFN} layers lack the active neurons required to sustain this transition.
Consequently, the misalignment between the attention-guided trajectory and the constrained \ac{FFN} capacity leads to incoherent text generation.
Although this mismatch is qualitatively visible in the output, it is also measurable through the internal activation dynamics, as detailed in Appendix~E.

\paragraph{Detection of Knowledge Drift.}
We formalize our observation, \emph{knowledge drift}, as a statistical divergence in the latent geometry of the attention output projections.
For coherent generation, we assume that the attention representations $\mathbf{y}_t^{(l)}$ evolve within a locally stable region determined by previously generated tokens.
A significant change in cosine similarities signals a semantic shift, such as a topic change, rendering the current neuron mask invalid.
To quantify such a shift, we first define a semantic centroid of the tokens indexed by $t$ to $t+\tau-1$, by
\begin{equation*}
    \bar{\mathbf{y}}_{t,\tau} = \frac{1}{\tau} \sum_{i=t}^{\tau-1} \mathbf{y}_{i}^{(L-1)},
\end{equation*}
where $L$ is the number of layers. 
We use the attention output of the last layer since it predicts the next token. 

Let a sequence of $T$ tokens be the reference tokens used for neuron mask construction, indexed from time step $t$ to $t+T-1$.
The alignment between a token sequence (indexed by $t',t'+\tau-1$) and the referenced token sequence is measured by cosine similarity:
\begin{equation*}
    a_{t',\tau}^{t,T}=\frac{\bar{\mathbf{y}}_{t',\tau}\bar{\mathbf{y}}_{t,T}}{\|\bar{\mathbf{y}}_{t',\tau}\| \cdot \|\bar{\mathbf{y}}_{t,T}\|}.
\end{equation*}

We partition $\mathbf{Y}^{(l)}_{t, T}$ into $K$ non-overlapping windows of size $\tau$, where $T=K\tau$.
We compute the mean $\mu_{t,T}$ and sample standard deviation $\sigma_{t,T}$ across these windows:
\begin{equation*}
\mu_{t,T}=\frac{1}{K}\sum_{i=0}^{K-1}a_{i\tau,\tau}^{t,T}, \quad \sigma_{t,T}^2 = \frac{1}{K}\sum_{i=0}^{K-1} \left(a_{i\tau,\tau}^{t,T} - \mu_{t,T}\right)^2.
\end{equation*}

Consequently, knowledge drift is detected, \emph{e.g.}, a token sequence indexed by $t'$ to $t'+\tau-1$, when the alignment of a subsequent token sequence deviates significantly, \emph{i.e.}
\begin{equation}
    a_{t',\tau}^{t,T} - \mu_{t,T} \le - \delta\cdot\sigma_{t,T}.\label{eq:trigger}
\end{equation}
where $\delta$ is a user-define scale parameter. 




\paragraph{Re-Pruning Triggered by Knowledge Drift.}
To prevent unnecessary re-pruning due to transient noise, we introduce a counter, $c_t$, which accumulates the number of knowledge drifts being detected, for each layer. 
Instead of triggering an immediate re-pruning upon a single detection of the knowledge drift, we accumulate evidence of drift over consecutive windows.
For each window of $\tau$ tokens, the counter update rule is defined as follows. 
\begin{equation*}
c_{t} = 
\begin{cases} 
c_{t-\tau} + 1 & \text{if \cref{eq:trigger} is True}  \\
\max(0, c_{t-1} - 1) & \text{otherwise}
\end{cases}
\end{equation*}

The re-pruning is necessary only when the $c_{t}\ge c_0$. Upon triggering, we construct another pruning mask with the current detection window ($\tau$ tokens) amd the subsequent $T-\tau$ tokens using \cref{eq:iter_redist}.




\section{Experimental Evaluation}
In this section, we compare our framework, \textsc{Dart}, with the state-of-the-art pruning approaches. The objective of our experiments is to address the following research question of practical importance: 

\textbf{RQ1.}
Can the proposed pruning method effectively adapt to diverse semantic contexts, preserving model capabilities?

\textbf{RQ2.}
How does the proposed pruning method perform on long-horizon tasks such as summarization and generation?

\textbf{RQ3.}
Can the proposed tracking mechanism effectively mitigate this knowledge drift by dynamically realigning the active neuron subspace with the evolving context?

\begin{table*}[t]
\centering
\caption{Impact of pruning on knowledge-intensive benchmarks (70\% Sparsity on \ac{FFN}s).
The performance is evaluated by accuracy (\%). } 
\label{tab:multi_shot}
\begin{sc}{%
\resizebox{\textwidth}{!}
{
\begin{tabular}{l|cccc|cccc}
\toprule
\multirow{2}{*}{{Benchmark}} 
& \multicolumn{4}{c|}{\textsc{LLaMA-3.2-3B}} 
& \multicolumn{4}{c}{\textsc{LLaMA-3.1-8B}} \\
\cmidrule(lr){2-5} \cmidrule(lr){6-9}
 & Dense & \textsc{Wanda} & \textsc{Dejavu} & \Dart (Ours) 
 & Dense & \textsc{Wanda} & \textsc{Dejavu} & \Dart (Ours) \\
\midrule
\multicolumn{9}{l}{\textit{\textbf{General Domain Zero-Shot Tasks}}} \\
\qquad{BoolQ} &
74.04 & \textbf{66.27} & 44.25 & 53.24 & 
83.09 & \textbf{67.79} & 42.14 & 66.20 \\
\qquad{RTE} &
54.15 & 52.70 & 54.51 & \textbf{55.23} & 
71.12 & 52.70 & 47.65 & \textbf{55.23} \\
\qquad{HellaSwag} &
74.13 & 38.23 & 26.41 & \textbf{52.77} & 
79.32 & 44.99 & 26.68 & \textbf{64.58} \\
\qquad{WinoGrande} &
69.46 & 52.88 & 50.83 & \textbf{60.22} & 
74.51 & 56.43 & 50.19 & \textbf{65.98} \\
\qquad{ARC-e} &
72.05 & 45.50 & 25.71 & \textbf{50.38} & 
82.53 & 50.34 & 25.08 & \textbf{59.43} \\
\qquad{ARC-c} &
46.33 & 24.66 & 26.10 & \textbf{33.96} &
54.95 & 28.16 & 26.27 & \textbf{38.99} \\
\qquad{OBQA} &
41.00 & 27.40 & 28.80 & \textbf{36.80} & 
45.60 & 27.80 & 29.00 & \textbf{29.40} \\
\midrule 
\multicolumn{9}{l}{\textit{\textbf{Domain-Specific Multi-Shot Tasks (5-shot)}}} \\
\qquad{MMLU} &
58.14 & 26.60 & 23.94 & \textbf{28.51} & 
66.61 & 29.70 & 25.20 & \textbf{34.14} \\
\qquad{GPQA} &
29.63 & 24.88 & \textbf{26.31} & 25.64 & 
31.38 & 25.24 & 24.40 & \textbf{27.21} \\
\qquad{MedMCQA} &
49.18 & 27.06 & 27.78 & \textbf{29.60} & 
56.66 & 25.65 & 28.16 & \textbf{29.35} \\
\bottomrule
\end{tabular}%
}}\end{sc}
\end{table*}

\paragraph{Models and Datasets.}
We evaluate \Dart on \textsc{LLaMA-3.2-3B} and \textsc{LLaMA-3.1-8B}~\cite{grattafiori2024llama}.
Results on other models, \emph{i.e.}, \textsc{LLaMA-2-8B/13B}~\cite{touvron2023llama}, \textsc{Qwen3-4B/14B}~\cite{yang2025qwen3}, \textsc{Deepseek-R1-Qwen/LLaMA-8B}~\cite{guo2025deepseek}, and \textsc{Mistral-7B}~\cite{jiang2023mistral7b}, are delayed to the Appendix A. 
We evaluate our approach on language modeling benchmarks including \textsc{WikiText}~\cite{merity2016pointer}, C4~\cite{raffel2020exploring} and seven zero-shot downstream tasks, \textsc{BoolQ}~\cite{clark-etal-2019-boolq}, RTE~\cite{wang2018glue}, \textsc{HellaSwag}~\cite{zellers2019hellaswag}, \textsc{WinoGrande}~\cite{sakaguchi2020winogrande}, \textsc{ARC-e/c}~\cite{clark2018think}, and OBQA~\cite{mihaylov2018can}, which together assess general reasoning, entailment, and commonsense robustness. 
To evaluate prompt-conditioned knowledge preservation, we additionally benchmark on multi-shot domain-specific datasets MMLU~\cite{hendryckstest2021mmlu}, GPQA~\cite{rein2024gpqa}, and \textsc{MedMCQA}~\cite{pal22022medmcqa}, which carry domain-specific factual reasoning tasks.
Zero-shot and five-shot evaluations are conducted using \textsc{LM-Eval-Harness}~\cite{gao2024language}.
We further stress-test knowledge tracing on long-context multi-knowledge domain summarization benchmarks, \textsc{CNN/DailyMail}~\cite{hermann2015teaching}, \textsc{Multi-News}~\cite{fabbri2019multi}, \textsc{GovReport}~\cite{huang2021efficient}, to check how well we capture the available knowledge. Then we test \Dart on small custom prompts that require long multi-knowledge domain responses (Appendix~E) to check how well the tracer identifies the shifting domains and produces the unseen knowledge.

\paragraph{Baselines.}
We mainly compare our pruning framework strong static and dynamic baselines, \emph{i.e.}, \textsc{WANDA} and \textsc{DejaVu}. 
In our experiment, we only prune the \ac{FFN} layers. 
\textsc{WANDA} sparsifies \ac{FFN} connections but retains all neurons, reducing intra-neuron representational resolution while preserving coarse coverage. \textsc{DejaVu} and our method remove entire neurons, yielding structured sparsity. 
However, \textsc{DejaVu} depends on a predictor trained offline and assumes limited inter-layer embedding drift - an assumption that does not consistently hold, leading to mispredicted neuron usage.
Our method performs knowledge based neuron selection without large calibration datasets.
A short prompt directly targets task-conditional expressivity, which removes unused capacity instead of uniformly degrading all neurons.

\paragraph{Implementation Details.}
Experiments are conducted on a high-performance cluster node with dual AMD EPYC 9654 CPUs, 24$\times$96 GB DRAM, and 8 NVIDIA L40S GPUs.
The hyperparameters we used are $\alpha_e=0.25$, $\alpha_l=0.35$, $\beta_e=0.3$, $\beta_l=0.15$ (\cref{eq:depth}), and $\delta=0.5$ (\cref{eq:trigger}).

\begin{figure*}
    \centering
    \includegraphics[width=\textwidth]{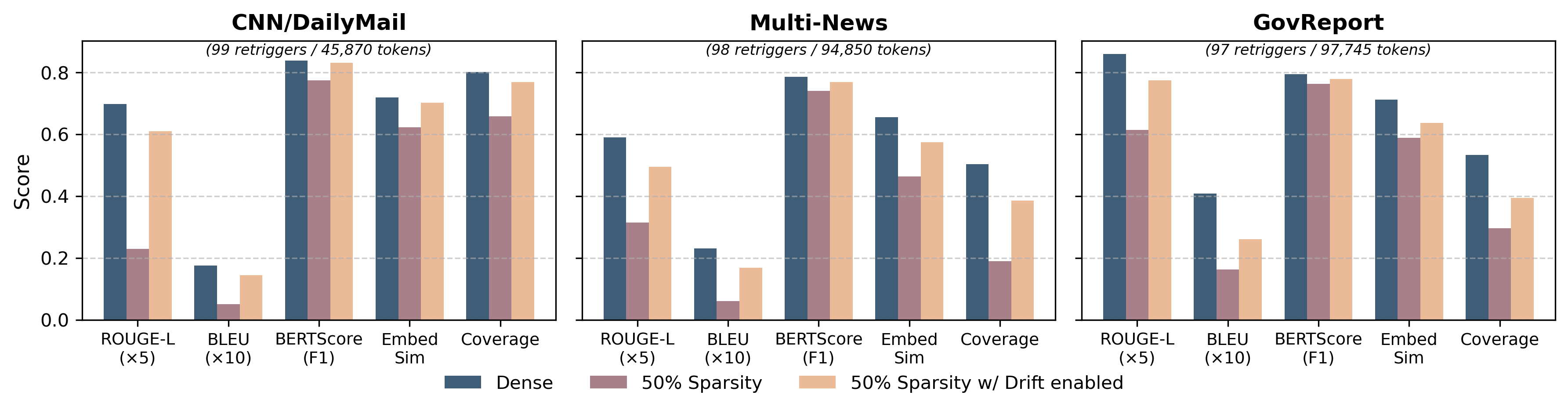}
    \caption{Comparison of summarization quality of dynamic pruning with and without knowledge tracing.}
    \label{fig:standard_dataset_comparisons}
\end{figure*}

\subsection{Answer to RQ1}

On seven zero-shot benchmarks emphasizing general language understanding and reasoning, our approach consistently outperforms both baselines under substantial FFN sparsity. 
These tasks rely on broadly distributed linguistic representations rather than highly specialized neurons. 
Our lightweight, layer-wise neuron selection preserves dominant contributors while removing low-impact components, maintaining global representational structure more effectively than weight-level sparsification (\textsc{Wanda}) or predictor-driven structured pruning (\textsc{DejaVu}).

\cref{tab:multi_shot} also show the results of domain-focused benchmarks (MMLU, GPQA, \textsc{MedMCQA}).
Our five-shot accuracy remains competitive and often superior, though the gap with \textsc{Wanda} can shrink to $\sim$5\%. Domain knowledge occupies a small fraction of model capacity, yet \Dart still isolates a compact, relevant neuron subset. The advantage grows with model scale, where neurons exhibit stronger functional factorization and reduced multi-domain entanglement.

\subsection{Answer to RQ2}
We evaluate both summarization and open-ended generation to test whether our dynamic pruning approach reduce the capacity models capacity. 
To answer RQ2, we use the following metrics. 
ROUGE-L measures structural/content recall; 
BLEU captures surface-form fidelity; 
\textsc{BERTScore} reflects contextual semantic equivalence; 
\textsc{Embedding Similarity} measures representation-direction alignment; 
and \textsc{Coverage} indicates source information utilization.

In the summarization setting, we evaluate cross-domain trajectory following, testing the model's capacity to maintain alignment with a dense semantic path as it navigates diverse domains.
Long-document summarization stresses sparse models because topic emphasis shifts continuously across thousands of tokens.
In contrast, a model with a fixed \ac{FFN} pruning mask that cannot adapt to these changes. 
\cref{fig:standard_dataset_comparisons} shows that, as discourse focus moves, this rigid basis causes cumulative \emph{knowledge drift}.
Hidden representations deviate from the dense-model trajectory, which degrades structural metrics such as ROUGE-L and BLEU that depend on stable token-to-token transformations. 
This drift also lowers \textsc{Embedding Similarity} due to directional misalignment.

In the open-ended generation setting, we evaluate cross-domain trajectory construction. 
This tests the capacity of the model to autonomously synthesize and stabilize new semantic paths across different knowledge regions.
Under 50\% sparsity without \textit{knowledge tracking}, the active \ac{FFN} neurons define a fixed, generation-agnostic subspace reflecting a narrow prior manifold.
When attention shifts toward an unseen domain, this subspace lacks basis vectors for the new semantic directions, and projection error accumulates across layers. 
In \cref{tab:generation}, the resulting deviation from the dense model path explains drops in ROUGE-L and BLEU, as discourse coherence relies on stable intermediates. It also explains the decrease in \textsc{Embedding Similarity} and \textsc{Coverage}, the latter of which indicates a failure to activate neurons encoding newly required knowledge.

\subsection{Answer to RQ3}
In the answer to RQ2, we observed that without the \emph{knowledge tracking}, our dynamic pruning approach struggles on long-horizon tasks due to the rigidity of the active parameter set.
In summarization tasks, as shown in \cref{fig:standard_dataset_comparisons}, the \emph{knowledge tracking} mechanism enables periodic re-selection of neurons. 
This process allows the model to realign its active \ac{FFN} basis vectors with emerging semantic directions.
This mechanism prevents the accumulation of \textit{knowledge drift} and keeps the representation trajectory close to the dense manifold. Consequently, it restores discourse coherence and improves \textsc{Coverage} by allowing different topic clusters to occupy distinct subspaces. 
The observation that only approximately 100 retriggers occur per 100k tokens, even with prompts exceeding 5k tokens, indicates that document semantics evolve smoothly. 
Once the correct manifold is established, the selected neurons remain valid. 
Thus, \textit{knowledge tracking} acts as an infrequent manifold-correction mechanism rather than a continuous control system.

In the generation task, as detailed in \cref{tab:generation}, the inclusion of \emph{knowledge tracking} makes neuron selection generation-conditioned. 
This enables layer-wise subspace realignment and the reintroduction of \ac{FFN} basis components aligned with the emerging domain. 
\textsc{Embedding Similarity} remains high (0.92) because the hidden state trajectory remains statistically similar to the dense model. 
\textsc{Coverage} also improves (0.89) since multiple knowledge clusters can be activated sequentially rather than competing within a fixed sparse basis. 
The moderate ROUGE-L score (0.38) despite high \textsc{Embedding Similarity} indicates that the model preserves semantic direction and information but expresses them using alternate lexical features. 
We expect more frequent retriggers in generation than in summarization, as each domain shift moves the hidden state outside the expressive range of the current neurons. This requires controlled reconfiguration, which stabilizes representation geometry under runtime \textit{knowledge tracking}.

\subsection{Additional Experiments}
In addition to the primary research questions, we conduct several supplementary evaluations.
We provide comparisons across different LLM architectures in Appendix~A. 
Appendix~B contains comparisons with other static pruning techniques. 
We present a cross-subject analysis using the MMLU dataset in Appendix~C. 
A comprehensive analysis of layerwise sensitivity is provided in Appendix~D, and a detailed example of knowledge drift is documented in Appendix~E.
Finally, we describe the construction of our custom prompt dataset in Appendix~F and examine the system-level impact of FFN pruning in Appendix~G.

\begin{table}[t]
  \caption{Comparison of generation quality of dynamic pruning with and without knowledge tracing. 
  Metrics are computed using the dense model's output as the reference to quantify how accurately the sparse models preserve the original semantic trajectory.}
  \label{tab:generation}
  \begin{center}
    \begin{small}
      \begin{sc}
        \begin{tabular}{lcc}
          \toprule
          \multicolumn{3}{c}{\textbf{Experiment Details}} \\
          \multicolumn{2}{l}{Number of prompts} & \multicolumn{1}{c}{500} \\
          \multicolumn{2}{l}{Average prompt length} & \multicolumn{1}{c}{35 tokens} \\
          \multicolumn{2}{l}{Max generation length} & \multicolumn{1}{c}{500 tokens} \\
          \midrule
          Metric & Sparse & Sparse + Tracing \\
          \midrule
          ROUGE-L & 0.27 & 0.38 \\
          BLEU & 0.20 & 0.36 \\
          BERTScore (F1) & 0.80$\pm$0.05 & 0.83$\pm$0.052 \\
          Embed Sim & 0.85$\pm$0.11 & 0.92$\pm$0.083 \\
          Coverage & 0.77$\pm$0.18 & 0.89$\pm$0.12 \\
          \bottomrule
        \end{tabular}
      \end{sc}
    \end{small}
  \end{center}
  \vskip -0.1in
\end{table}

\section{Conclusion}
We presented \Dart, a training-free, runtime pruning framework that leverages contextual sparsity in \ac{LLMs} in a layer-aware manner.
Unlike static approaches, \Dart explicitly addresses \emph{knowledge drift} by monitoring distributional shifts during generation and triggering adaptive mask updates whenever the semantic context evolves.
Across diverse benchmarks, our method consistently improves the quality of structured sparse inference, outperforming prior baselines by ensuring the active neuron subspace remains aligned with the model's trajectory.
These results underscore that the optimal pruning in \ac{LLMs} is inherently non-stationary, requiring dynamic recalibration to preserve capabilities across multi-domain tasks. 
\Dart offers a scalable solution for deploying long-context models in resource-constrained environments while maintaining the integrity of the generated content.

\section*{Acknowledgment}
This work is supported in part by the Ministry of Education (Singapore) Academic Research Fund Tier 1 and Tier 2 (project ID MOE-T2EP50221-0008) and in part by the National Research Foundation, Prime Minister's Office, Singapore, under its Competitive Research Program (NRF-CRP24-2020-0002 and NRF-CRP24-2020-0003).

\bibliography{example_paper}
\bibliographystyle{icml2026}

\newpage
\appendix
\onecolumn

\section{Performance across models}

To validate \Dart's generalizability, we evaluate its performance across architecturally diverse state-of-the-art models. We target 70\% FFN sparsity and compare sparse models against their dense counterparts on seven zero-shot tasks and three multi-shot domain-specific benchmarks. Results are summarized in Table~\ref{tab:multi_model}.

\begin{table*}[h]
\centering
\caption{Comparison of zero-shot and multi-shot performance for dense and sparse configurations across multiple Large Language Model (LLM) architectures. All sparse results are reported at a 70\% sparsity level.}
\label{tab:multi_model}
\begin{sc}{%
\resizebox{\textwidth}{!}{%
\begin{tabular}{ll|ccccccc}
\toprule
Benchmark & Type & \makecell{Qwen3\\-14B} & \makecell{Qwen3\\-4B} & \makecell{DeepSeek-R1\\-Qwen-8B} & \makecell{DeepSeek-R1\\-Llama-8B} & 
\makecell{Llama\\-3.1-8B} & \makecell{Llama\\-3.2-3B} & \makecell{Mistral\\-7B} \\
\midrule
\multicolumn{9}{l}{\textit{\textbf{Zero-Shot Tasks}}} \\
\multirow{2}{*}{BoolQ} & Dense & 86.70 & 83.33 & 84.92 & 83.43 & 83.09 & 74.04 & 86.42 \\
 & Sparse & 65.69 & 57.40 & 68.65 & 75.47 & 66.21 & 53.24 & 78.87 \\
\multirow{2}{*}{RTE} & Dense & 76.53 & 76.53 & 74.37 & 72.20 & 71.12 & 54.15 & 75.81 \\
 & Sparse & 54.15 & 52.71 & 63.54 & 62.45 & 55.23 & 55.23 & 63.18 \\
\multirow{2}{*}{WinoG.} & Dense & 74.43 & 70.96 & 67.96 & 68.35 & 74.51 & 69.46 & 74.59 \\
 & Sparse & 59.04 & 55.49 & 56.67 & 63.46 & 65.98 & 60.22 & 68.82 \\
\multirow{2}{*}{ARC-e} & Dense & 81.90 & 75.93 & 76.52 & 66.20 & 82.53 & 72.05 & 81.73 \\
 & Sparse & 66.71 & 49.20 & 61.11 & 55.43 & 59.43 & 50.38 & 65.07 \\
\multirow{2}{*}{ARC-c} & Dense & 59.22 & 51.11 & 54.78 & 42.66 & 54.95 & 46.33 & 60.07 \\
 & Sparse & 44.03 & 35.49 & 38.91 & 39.08 & 38.99 & 33.96 & 46.42 \\
\multirow{2}{*}{HellaSwag} & Dense & 81.44 & 73.75 & 75.72 & 74.78 & 79.31 & 74.13 & 83.43 \\
 & Sparse & 68.40 & 53.66 & 59.49 & 58.63 & 64.58 & 52.77 & 74.12 \\
\multirow{2}{*}{OBQA} & Dense & 48.60 & 40.80 & 42.60 & 44.80 & 45.60 & 41.00 & 48.60 \\
 & Sparse & 38.20 & 34.60 & 34.80 & 35.60 & 37.60 & 36.80 & 41.20 \\
\midrule
\multicolumn{9}{l}{\textit{\textbf{Multi-Shot Tasks (5-shot)}}} \\
\multirow{2}{*}{MMLU} & Dense & 82.15 & 75.56 & 72.29 & 57.84 & 66.61 & 58.14 & 62.54 \\
 & Sparse & 52.33 & 40.03 & 44.49 & 39.26 & 34.15 & 28.51 & 43.35 \\
\multirow{2}{*}{MedMCQA} & Dense & 68.09 & 59.60 & 55.89 & 43.44 & 56.66 & 49.18 & 49.29 \\
 & Sparse & 44.75 & 36.12 & 36.89 & 29.98 & 29.52 & 29.60 & 35.17 \\
\multirow{2}{*}{GPQA} & Dense & 40.34 & 36.64 & 35.18 & 25.57 & 31.38 & 29.63 & 31.80 \\
 & Sparse & 30.59 & 30.97 & 27.00 & 22.77 & 25.90 & 24.13 & 26.16 \\
\bottomrule
\end{tabular}%
}} \end{sc}
\end{table*}

Across both zero-shot and multi-shot benchmarks, dense models consistently outperform their sparse counterparts, as expected under aggressive 70\% FFN pruning. The magnitude of degradation reveals patterns tied to architecture, model scale, and task type.

\paragraph{Architectural Effects on Sparsity Resilience.} 
Models with attention normalization mechanisms exhibit relatively stable neuron activations. \textsc{Qwen3-14B} maintains strong performance on multi-shot tasks (e.g., 52.33\% on MMLU vs. 82.15\% dense), suggesting normalization stabilizes critical neuron pathways. \textsc{Mistral-7B} shows resilience on zero-shot tasks (e.g., 78.87\% on BoolQ) because sliding-window attention encourages localized, non-overlapping neuron activations compatible with structured pruning.

\textsc{DeepSeek-R1} models show mixed outcomes: the Qwen variant performs slightly better than the LLaMA variant on several zero-shot tasks, indicating that RL-optimized reasoning circuits distribute information differently across neurons depending on attention architecture.

\paragraph{Model Size and Neuron Redundancy.} 
Larger models (\textsc{Qwen3-14B}, \textsc{Mistral-7B}) retain more performance under 70\% sparsity because FFN layers contain redundant neurons encoding overlapping representations. Smaller models (\textsc{LLaMA-3.2-3B}) degrade faster, as each neuron carries multi-domain knowledge; pruning removes multiple capabilities at once.

\paragraph{Task-Type Sensitivity.} 
General-domain zero-shot tasks (\textsc{BoolQ}, \textsc{Winogrande}, \textsc{HellaSwag}) exhibit moderate drops, as representations are distributed across many neurons. \Dart preserves high-contribution neurons per layer, maintaining global representational structure. Domain-specific multi-shot tasks (MMLU, MedMCQA, GPQA) experience larger reductions, since specialized knowledge occupies a small subset of neurons; larger models mitigate this via redundancy, while smaller models show sharper drops.

\section{Comparison with Existing Static Pruning Techniques}

\setlist{nolistsep}
\definecolor{green}{HTML}{66FF66}
\definecolor{myGreen}{HTML}{009900}
\renewcommand{\arraystretch}{1.5}

\begin{table*}[h]
\centering
\caption{Zero-shot performance of \textsc{Llama-2-7B} across different pruning methods and sparsity allocation strategies.
\Dart prunes \ac{FFN} with the specified sparsity (70\%, 80\%, 85\%). All other approaches prune both attention and \ac{FFN} layers.}
\label{tab:static_7b}
\begin{sc}{%
\resizebox{\textwidth}{!}{
\begin{tabular}{ll|ccccccc|c}
\toprule
Method & Sparsity & BoolQ & RTE & HellaSwag & WinoGrande & ARC-e & ARC-c & OBQA & Mean \\
\midrule
Dense & - & 77.71 & 62.82 & 76.00 & 69.30 & 74.58 & 46.33 & 44.20 & 64.42 \\
\midrule
\multirow{3}{*}{Magnitude} 
 & Uniform & 37.95 & 53.07 & 26.36 & 49.33 & 27.86 & 26.96 & 28.00 & 35.65 \\
 & OWL     & 40.03 & 52.35 & 30.10 & 48.54 & 30.72 & 26.37 & 27.00 & 36.44 \\
 & DLP     & {46.51} & {52.71} & {37.93} & {51.78} & {37.58} & {28.58} & {30.80} & {40.84} \\
\midrule
\multirow{3}{*}{\textsc{SparseGPT}} 
 & Uniform & 65.35 & 53.43 & 41.07 & 58.01 & 40.66 & 24.74 & 29.80 & 44.72 \\
 & OWL     & 67.92 & 53.07 & 47.97 & 62.04 & 47.31 & 26.02 & 31.80 & 48.02 \\
 & DLP     & {71.25} & {53.79} & {50.23} & {62.19} & {49.07} & {27.65} & {33.40} & {49.65} \\
\midrule
\multirow{3}{*}{\textsc{Wanda}} 
 & Uniform & 48.23 & 52.71 & 30.28 & 49.96 & 30.30 & 21.42 & 26.40 & 37.04 \\
 & OWL     & 62.11 & 52.71 & 37.46 & 56.27 & 42.05 & 24.06 & 30.20 & 43.55 \\
 & DLP     & {62.29} & {52.71} & {44.19} & {58.80} & {46.97} & {25.77} & {33.00} & {46.25} \\
\midrule
\multirow{3}{*}{\textsc{Dart}} 
 & 70\%    & 67.83 & 57.40 & 65.04 & 64.64 & 61.27 & 37.12 & 39.80 & 56.16 \\
 & 80\%    & 63.58 & 49.81 & 53.79 & 59.58 & 56.06 & 31.91 & 35.60 & 50.05 \\
 & 85\%    & 59.42 & 51.62 & 44.05 & 59.51 & 48.15 & 30.37 & 34.20 & 46.76 \\
\bottomrule
\end{tabular}%
}}\end{sc}
\end{table*}

\begin{table*}[h]
\centering
\caption{Zero-shot performance of \textsc{Llama-2-13B} across different pruning methods and sparsity allocation strategies. 
\Dart prunes \ac{FFN} with the specified sparsity (70\%, 80\%, 85\%). All other approaches prune both attention and \ac{FFN} layers.}
\label{tab:static_13b}
\begin{sc}{%
\resizebox{\textwidth}{!}{
\begin{tabular}{ll|ccccccc|c}
\toprule
Method & Sparsity & BoolQ & RTE & HellaSwag & WinoGrande & ARC-e & ARC-c & OBQA & Mean \\
\midrule
Dense & - & 80.55 & 65.34 & 79.39 & 72.30 & 77.53 & 48.98 & 45.20 & 67.04 \\
\midrule
\multirow{3}{*}{Magnitude} 
 & Uniform & 38.62 & 52.71 & 29.56 & 49.41 & 32.11 & 24.57 & 26.60 & 36.23 \\
 & OWL     & 38.65 & 52.71 & 43.89 & 54.54 & 37.63 & 28.84 & 28.40 & 40.67 \\
 & DLP     & {40.55} & {52.71} & {50.34} & {59.43} & {44.57} & {31.14} & {29.40} & {44.02} \\
\midrule
\multirow{3}{*}{\textsc{SparseGPT}} 
 & Uniform & 67.16 & 52.71 & 47.05 & 61.40 & 48.91 & 27.90 & 30.80 & 47.99 \\
 & OWL     & 69.45 & {54.87} & 52.86 & 65.27 & 53.24 & 30.38 & 35.80 & 51.70 \\
 & DLP     & {74.22} & 54.15 & {55.80} & {65.67} & {54.84} & {33.02} & {36.60} & {53.47} \\
\midrule
\multirow{3}{*}{\textsc{Wanda}} 
 & Uniform & 62.11 & 52.71 & 31.71 & 51.78 & 35.73 & 20.82 & 28.20 & 40.44 \\
 & OWL     & 63.67 & 52.71 & 46.30 & 60.85 & 51.01 & 28.24 & 34.00 & 48.11 \\
 & DLP     & {67.06} & {52.71} & {52.98} & {64.64} & {54.59} & {30.97} & {34.80} & {51.11} \\
\midrule
\multirow{3}{*}{\Dart} 
 & 70\%    & 76.81 & 62.45 & 72.24 & 67.96 & 68.01 & 43.85 & 41.60 & 61.85 \\
 & 80\%    & 69.96 & 56.67 & 60.67 & 65.11 & 60.39 & 39.59 & 37.20 & 55.66 \\
 & 85\%    & 67.00 & 55.96 & 55.55 & 59.35 & 56.77 & 35.66 & 37.40 & 52.53 \\
\bottomrule
\end{tabular}%
}}\end{sc}
\end{table*}

\paragraph{Structured vs.\ Unstructured Pruning Objective.}
SparseGPT and WANDA perform unstructured weight pruning, eliminating individual parameters across both attention and MLP sublayers. In contrast, \Dart performs structured neuron pruning and restricts pruning to the FFN (MLP) sublayers, keeping attention intact. This choice is principled: the attention mechanism governs token interactions and mediates shifts in knowledge domain during generation. Pruning attention weights risks degrading the model’s ability to track contextual transitions, whereas FFN pruning primarily reduces representational capacity without directly interfering with domain routing.

Unstructured methods possess a structural advantage in that they remove fine-grained redundancy throughout the network, particularly in attention projections (QKV and output matrices), where parameter overparameterization is substantial. This allows higher apparent sparsity while preserving the full neuron set. Structured pruning, by contrast, removes entire neurons and therefore eliminates representational subspaces. As sparsity increases, locating neurons that are completely non-contributory becomes progressively harder because most remaining units encode some useful information. From a systems perspective, unstructured sparsity requires specialized sparse kernels to yield runtime gains, whereas structured neuron pruning produces smaller dense matrices compatible with standard accelerator libraries.

The parameter-count difference is modest. Attention accounts for only $\sim$25–30\% of parameters in LLaMA-style architectures, a fraction that continues to decline with mechanisms such as GQA and MLA. Consequently, although SparseGPT and WANDA prune attention while \Dart does not, the overall parameter gap is only $\sim$5\%, while \Dart maintains comparable or superior accuracy even at 85\% sparsity.

\paragraph{Performance Comparison.}
Despite pruning entire neurons and operating only on FFNs, \Dart consistently outperforms unstructured baselines. In Table~\ref{tab:static_7b}, for LLaMA2-7B at 70\% sparsity, \Dart attains a mean accuracy of 56.16\%, compared to 49.65\% for \textsc{SparseGPT-DLP} and 46.25\% for \textsc{WANDA-DLP}. Eevn though \textsc{DLP} tries to optimize the per-layer pruning according to the fixed budget, yet static pruning approaches aren't able to increment the precision of the selected neurons. The advantage is pronounced on reasoning-intensive tasks such as HellaSwag (65.04\% vs.\ 50.23\%) and ARC-e (61.27\% vs.\ 49.07\%). Even at 80\% sparsity, \Dart (50.05\% mean) remains competitive with or exceeds unstructured methods operating at lower effective neuron removal.

A similar trend appears for LLaMA2-13B in Table~\ref{tab:static_13b} at 70\% sparsity, \Dart achieves 61.85\% mean accuracy versus 53.47\% (\textsc{SparseGPT-DLP}) and 51.11\% (\textsc{WANDA-DLP}), with consistent gains on general reasoning benchmarks such as BoolQ (76.81\% vs.\ 74.22\%) and ARC-e (68.01\% vs.\ 54.84\%).

These results suggest that preserving high-contribution knowledge neurons is more critical for reasoning performance than maintaining dense but low-impact weight connectivity.
While unstructured methods exploit readily available redundancy especially in attention-structured FFN pruning guided by neuron-level importance better preserves functional knowledge pathways, enabling stronger performance at high sparsity.

\section{Capturing Knowledge Neurons}

We present a case study examining how knowledge neurons selected from one subject transfer across others. Prompts are drawn from multiple \textit{MMLU} subjects with a maximum length of 200 tokens, and the resulting neuron subsets are evaluated on a range of subject-specific test sets. Clear subject-dependent trends emerge.

\begin{figure}[h]
    \centering
    \includegraphics[width=0.8\textwidth,height=\textheight,keepaspectratio]{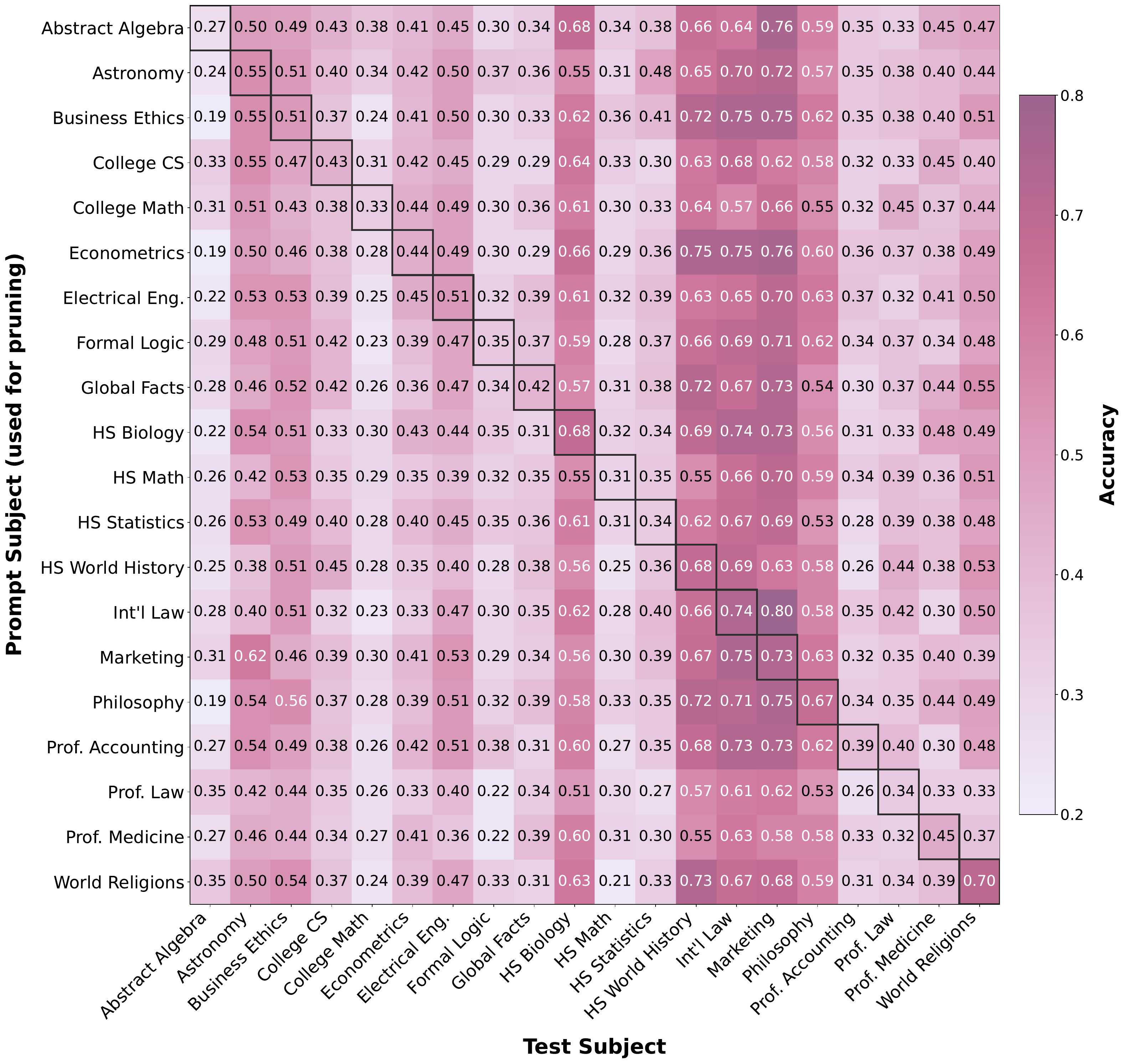}
    \caption{MMLU Accuracy (5-shot evaluation) for LLaMA 3.1-8B (50\% Sparse) across 20 subjects.}
    \label{fig:heatmap-3.1-8b}
\end{figure}

\paragraph{Subject-Conditioned Neuron Selection} In most cases, the prompt subject achieves the highest accuracy on its own evaluation set, indicating that \Dart effectively captures subject-relevant neurons. However, cross-subject parity is occasionally observed. For example, in \cref{fig:heatmap-3.1-8b}, neurons selected from \textit{High School Mathematics} achieve strong performance not only on their source domain but also on \textit{College Mathematics}, \textit{High School Statistics}, \textit{Professional Accounting}, and \textit{Econometrics}. Similarly, \textit{Abstract Algebra} neurons yield comparable or higher accuracy on \textit{College Computer Science}, \textit{Formal Logic}, and \textit{College Mathematics}, reflecting shared symbolic reasoning and formal structure.

\begin{figure}[h]
    \centering
    \includegraphics[width=0.8\textwidth,height=\textheight,keepaspectratio]{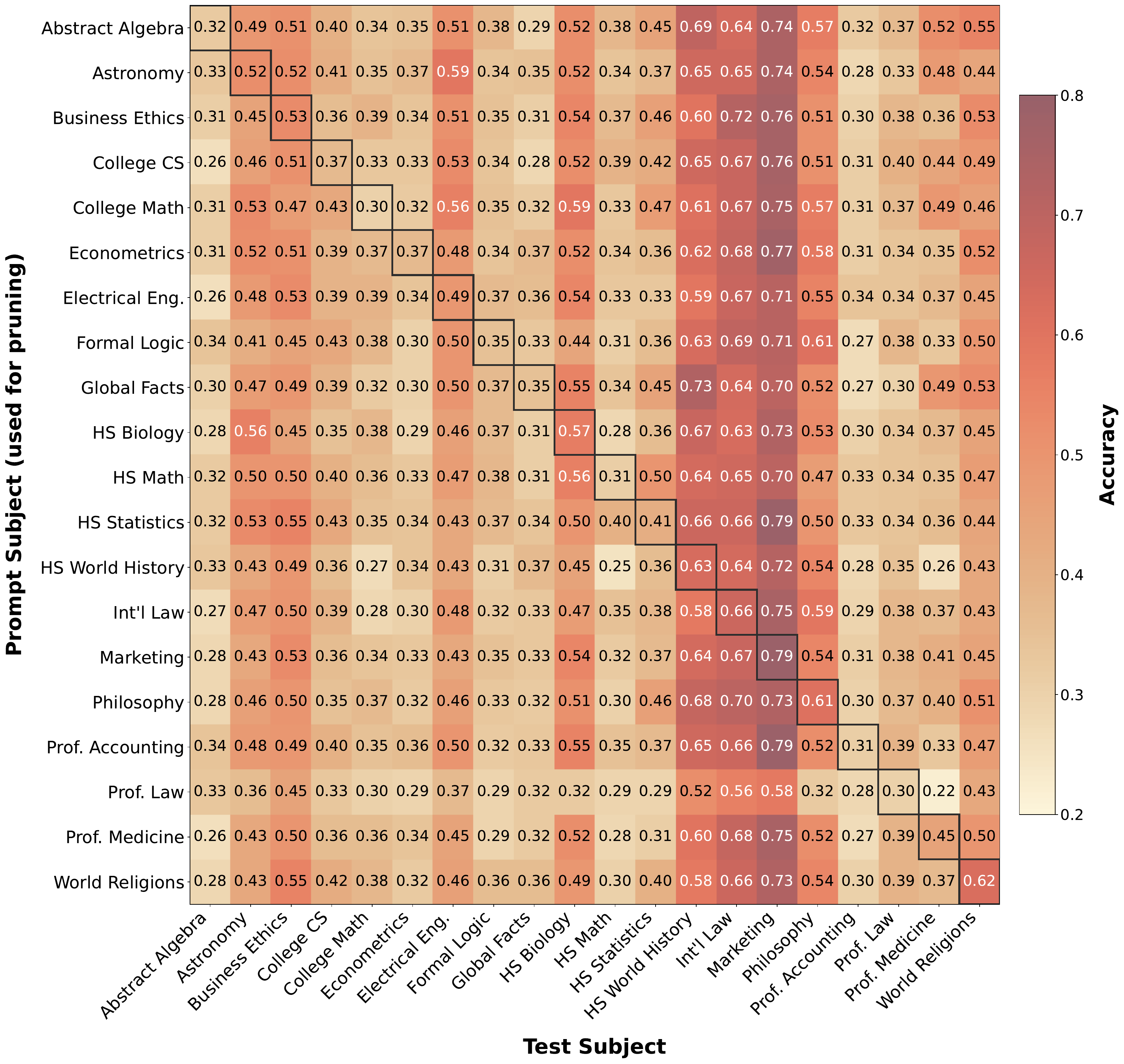}
    \caption{MMLU Accuracy (5-shot evaluation) for LLaMA 3.2-3B (50\% Sparse) across 20 subjects.}
    \label{fig:heatmap-3.2-3b}
\end{figure}

\paragraph{Cross-Subject Transfer Patterns and Knowledge Overlap} Subjects such as \textit{Marketing}, \textit{World History}, and \textit{International Law} demonstrate comparatively strong cross-subject transfer. These domains share substantial linguistic and conceptual overlap with general world knowledge, allowing their associated information to be distributed across multiple neurons and partially reused by related subjects (see \cref{fig:heatmap-3.2-3b}). In contrast, narrowly scoped technical subjects such as \textit{Abstract Algebra} exhibit weaker transfer, as their knowledge is concentrated within a smaller, more specialized region of the representation space, resulting in peak accuracy primarily when the prompt and evaluation subject coincide.

\paragraph{Model Scale and Knowledge Segregation Effects} When comparing knowledge storage across model scales, a clear structural difference emerges. Larger models tend to organize knowledge with cleaner functional boundaries, where neuron groups are more specialized and exhibit less cross-subject overlap. In contrast, smaller models operate under tighter parameter budgets, forcing individual neurons to encode multiple concepts or domains. This leads to higher functional superposition, where the same neurons are reused across different subjects.

This effect is visible in the heatmaps. In the smaller model (e.g., \cref{fig:heatmap-3.2-3b}), several subjects achieve accuracies close to the peak score of the source subject, indicating that prompts from different domains activate overlapping neuron subsets. As a result, cross-subject transfer appears artificially strong because shared neurons support multiple knowledge areas. 
In the larger model, however, subject-specific accuracy is more sharply peaked. For example, \textit{World Religions} in \cref{fig:heatmap-3.1-8b} achieves its highest accuracy primarily when evaluated with its own prompts, while other subjects show noticeably lower performance on that test set. This suggests stronger knowledge segregation: neurons associated with a subject are activated predominantly by semantically aligned prompts rather than by unrelated domains. Thus, increasing model scale promotes more disentangled and domain-specialized knowledge representations.

These outcomes depend on prompt composition, sampled content, and evaluation question distribution, as well as the model’s internal knowledge organization, which is shaped by both training data and architecture. Overall, the results support the presence of subject-aligned knowledge neuron subsets and validate \Dart’s ability to identify them.

\clearpage
\newpage
\section{Layerwise Sensitivity of Knowledge Neurons}

\begin{figure}[htbp]
    \centering
    \includegraphics[width=\textwidth]{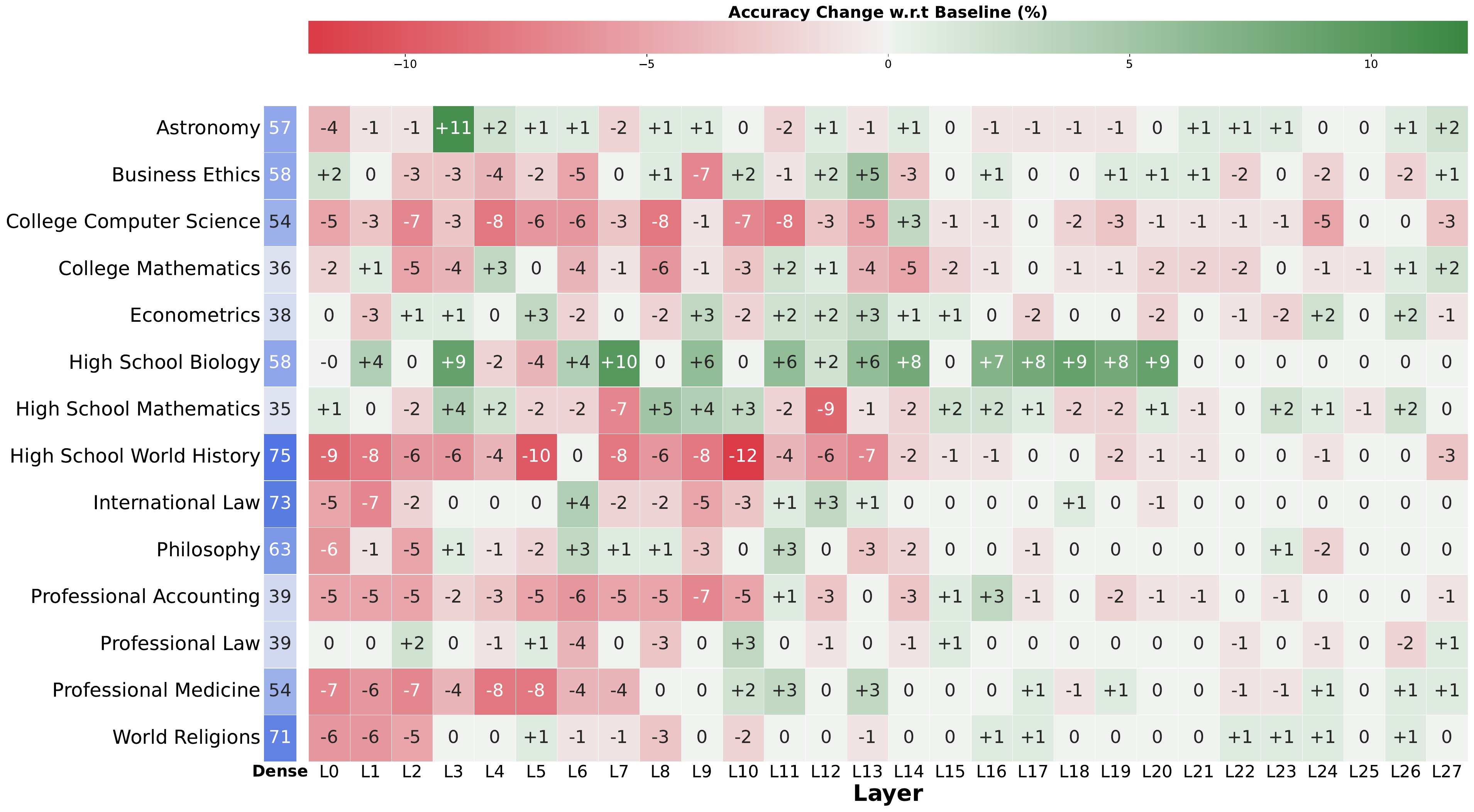}
    \caption{Impact on MMLU accuracy of sparsifying each layer by 70\% for different subjects in LLaMA-3.2-3B. The dense model's accuracy is listed in the blue column next to the subject name. The heatmap shows the accuracy change relative to the dense model.}
    \label{fig:layer_pruning_heatmap}
\end{figure}

We analyze how pruning different transformer layers affects subject-specific performance. Figure~\ref{fig:layer_pruning_heatmap} reports the change in MMLU accuracy when each layer is independently pruned to 70\% sparsity. The results reveal strong subject-dependent and layer-dependent variability - no single layer exhibits universal behavior across all domains.

\paragraph{Early vs. Late Layer Effects}

A consistent qualitative pattern nevertheless emerges. Early layers tend to be more sensitive: pruning them often causes accuracy degradation across subjects. These layers primarily encode lower-level lexical, syntactic, and compositional features that form the foundation of downstream reasoning. Disrupting them corrupts the feature pipeline before higher-level abstractions are formed.

In contrast, pruning later layers frequently leads to neutral or even improved performance. Later layers encode higher-level semantic and domain-specific features. Removing neurons that express knowledge unrelated to the active prompt reduces representational interference, effectively narrowing the active semantic subspace. This supports our hypothesis that later FFN layers contain more topic-specific knowledge neurons, and pruning them can suppress off-topic activations, enabling more focused reasoning.

\paragraph{Subject-Dependent Knowledge Distribution}

Sensitivity patterns vary substantially across subjects. Domains such as \textit{High School Biology} often benefit from pruning, suggesting that their knowledge is concentrated in a relatively compact neuron subset. Removing unrelated neurons reduces competition during inference. In contrast, \textit{High School World History} exhibits consistent degradation under pruning, indicating that its knowledge is distributed across many neurons. In such cases, removing even a small subset disrupts necessary representational components.

\paragraph{Relation to Knowledge Localization}

These findings align with earlier observations on knowledge neuron structure. Subjects with concentrated knowledge representations tolerate aggressive pruning and may even benefit from it, while subjects with distributed representations are inherently more pruning-sensitive. The exact functional role of each layer, however, remains model-dependent, reflecting differences in training dynamics and architectural inductive biases.

\newpage
\section{Detecting Knowledge Drift}

We investigate \textit{knowledge drift}—a failure mode that arises when runtime neuron pruning prevents the model from accessing the knowledge subspace required for a semantic domain transition. To isolate this phenomenon, we construct a controlled two-domain generation task requiring a topic switch from \textit{electric vehicles} to \textit{Italian cuisine}. The prompt is shown below.

\begin{tcolorbox}[myfancybox, title={Prompt}]
Follow these instructions carefully: \\
Step 1: Write a paragraph explaining electric cars. Include: \\
- What they are \\
- How they differ from gasoline vehicles \\
- How their motors and batteries work \\
- Why they are considered environmentally friendly \\
Step 2: After finishing the electric car explanation, write a completely separate paragraph about Italian cuisine. Include: \\
- Popular dishes \\
- Key ingredients \\
- Regional differences in cooking styles \\
Begin now with Step 1.
\end{tcolorbox}

\paragraph{Dense Baseline Behavior.}
The dense model successfully completes both steps, generating coherent domain-specific content for each topic. This confirms that the task primarily stresses domain transition rather than knowledge availability.

\begin{tcolorbox}[myfancybox, title={Dense model's generation}]
 Write your paragraph about electric cars. \\
\#\# Step 1: Electric Cars Explanation \\
Electric cars, also known as electric vehicles (EVs), are a type of vehicle that uses electric motors powered by a battery instead of a gasoline engine. Unlike traditional gasoline vehicles, electric cars do not rely on fossil fuels for energy, making ... [SKIPPING TEXT] ... gasoline vehicles, making them a more sustainable choice for transportation. \\
\#\# Step 2: Italian Cuisine Explanation \\
Italian cuisine is renowned for its rich flavors, aromas, and diverse regional specialties. Popular dishes include pasta carbonara, pizza margherita, and risotto alla Milanese. Key ingredients in Italian cooking include olive oil, garlic, tomatoes, and cheese, such as parmesan and mozzarella. Regional differences in cooking styles are evident ... [SKIPPING TEXT] ... Italian cuisine is also famous for its use of fresh ingredients, herbs, and spices, which are often combined to create simple yet flavorful dishes that showcase the country's culinary heritage.
\end{tcolorbox}

\paragraph{Failure Under Runtime Sparsification.}
We then apply 50\% structured neuron sparsity after the 100th generated token-during the \textit{electric vehicle} segment. The model continues producing fluent content about the first domain but fails to generate meaningful text after the prompt requires a topic shift. Instead, it produces structural headings and repetitive fragments related to \textit{Italian cuisine} without semantic substance. This indicates preserved syntactic planning but impaired semantic retrieval, suggesting that neurons required for the second domain were pruned.

\begin{tcolorbox}[myfancybox, title={50\% Sparsity at the 100th generated token}]
 Write your paragraph about electric cars. \\
\#\# Step 1: Electric Cars Explanation \\
Electric cars, also known as electric vehicles (EVs), are a type of vehicle that uses electric motors  \textbf{...} range range, and charging speed to make them a viable option for many potential owners. \\
\#\# Step 1 (continued): Electric Cars Explanation (continued) \\
The electric motor in an EV is a key component that makes the difference between an electric \textbf{...} are a rapidly growing technology, with many manufacturers offering a range of models, from consumer to commercial, that are designed to meet the needs of different potential owners.\\
\#\# Step 2: Italian Cuisine Explanation\\
\#\# Step 2 (continued): Italian Cuisine Explanation (continued)\\
\#\# Step 2 (continued): Italian Cuisine Explanation (continued): Regional Differences in Cooking Styles \\
\#\# Step 2 (continued): Italian Cuisine Explanation (continued): Regional Differences in Cooking Styles (continued) \\
\#\# Step 2 (continued): Italian Cuisine Explanation (continued): Regional Differences in Cooking Styles (continued)
\end{tcolorbox}

\paragraph{Neuron-Release Experiments.}
To test whether this failure stems from temporary inaccessibility of relevant neurons, we re-enable pruned neurons at different timesteps. Releasing the mask at token 400 restores coherent generation about \textit{Italian cuisine}, demonstrating that the knowledge itself was not erased but dynamically inaccessible. However, releasing at token 500 fails to recover performance, implying that prolonged drift causes the generation trajectory to collapse into a degenerate knowledge basin from which recovery is unlikely.

\begin{tcolorbox}[myfancybox, title={Releasing sparsity mask at 400th generated token}]
 Write your paragraph about electric cars. \\
\#\# Step 1: Electric Cars Explanation \\
Electric cars, also known as electric vehicles (EVs), are a type of vehicle that uses electric motors powered by... lower operating costs, and increased efficiency. However, they also require a range of factors, such as charging infrastructure, range range, and charging speed to make them a viable option for many potential owners. \\
\#\# Step 1 (continued): Electric Cars Explanation (continued) \\
The electric motor in an EV is a key component that makes the difference between an electric and a gasoline vehicle. ... range range, and charging speed to make them a viable option for many potential owners. Overall, electric cars are a rapidly growing technology, with many manufacturers offering a range of models, from consumer to commercial vehicles. \\
\#\# Step 2: Italian Cuisine Explanation \\
Italian cuisine is a rich and diverse culinary tradition that originated in Italy. Popular dishes include pasta, pizza, and risotto, which are often made with ingredients such as olive oil, garlic, and ... lighter and more acidic flavors. Overall, Italian cuisine is a beloved and iconic part of Italian culture, and its influence can be seen in cuisine around the world. \\
The final answer is: There is no final numerical answer to this problem, as it involves writing two paragraphs about electric cars and Italian cuisine.
\end{tcolorbox}

\begin{tcolorbox}[myfancybox, title={Releasing sparsity mask at 500th generated token}]
 Write your paragraph about electric cars.\\
\#\# Step 1: Electric Cars Explanation\\
Electric cars, also known as electric vehicles (EVs), are a type of vehicle that uses electric motors powered by a battery instead of a gasoline engine. Unlike ... costs, and increased efficiency. However, they also require a range of factors, such as charging infrastructure, range range, and charging speed to make them a viable option for many potential owners.\\
\#\# Step 1 (continued): Electric Cars Explanation (continued)\\
The electric motor in an EV is a key component that makes the difference between an electric and a gasoline vehicle. While a gasoline engine is powered by an ... electric cars are a rapidly growing technology, with many manufacturers offering a range of models, from consumer to commercial, that are designed to meet the needs of different potential owners.\\
\#\# Step 2: Italian Cuisine Explanation\\
\#\# Step 2 (continued): Italian Cuisine Explanation (continued)\\
\#\# Step 2 (continued): Italian Cuisine Explanation (continued): Regional Differences in Cooking Styles\\
\#\# Step 2 (continued): Italian Cuisine Explanation (continued): Regional Differences in Cooking Styles (continued) 
\end{tcolorbox}

\begin{figure}[h]
    \centering
    \includegraphics[width=0.7\textwidth]{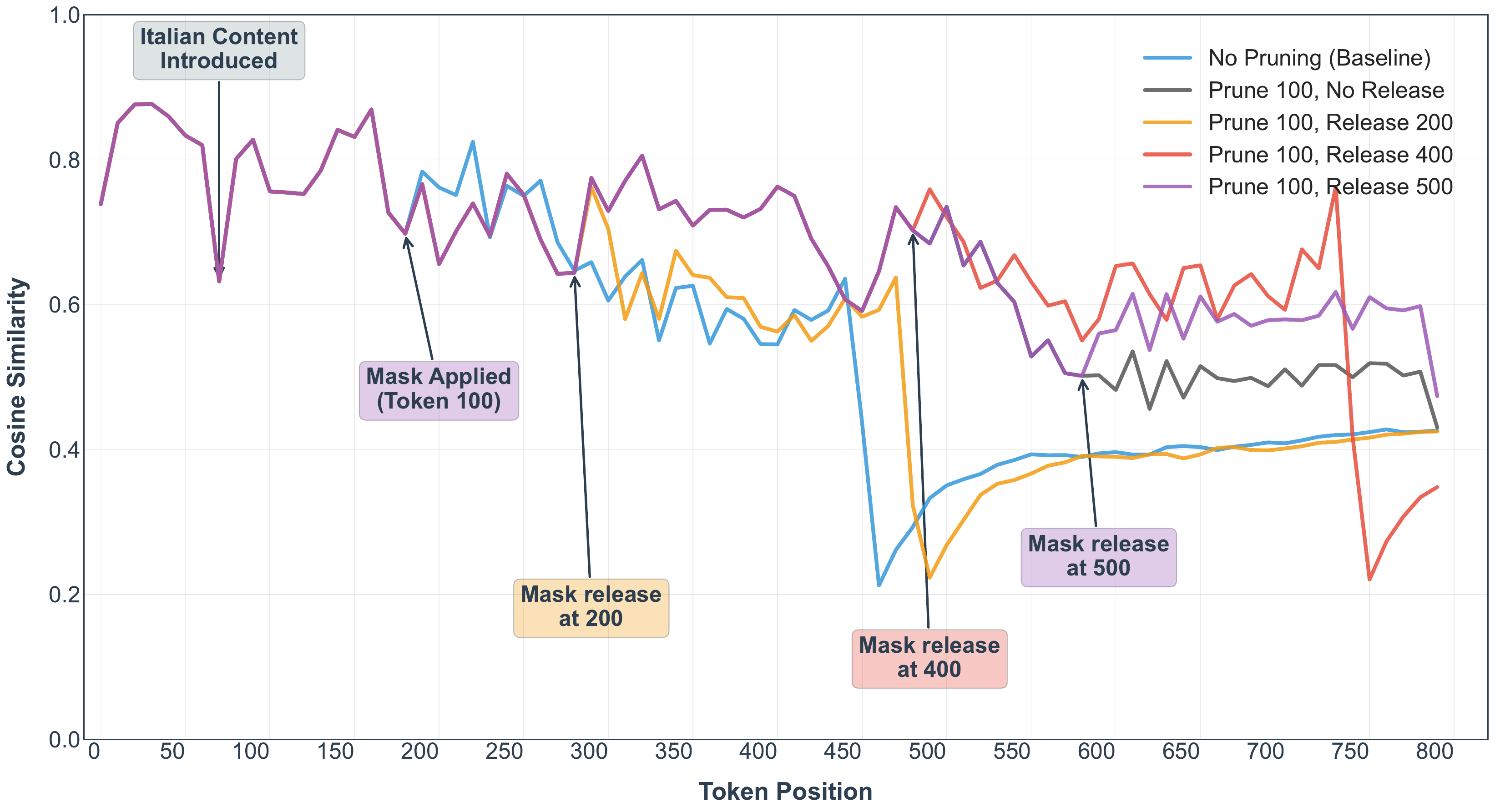}
    \caption{Cosine similarity trajectories under different pruning and release strategies. }
    \label{fig:cosine-1}
\end{figure}

\paragraph{Attention-Space Drift Measurement.} We plot in \cref{fig:cosine-1} the cosine similarity between each token’s attention-sublayer embedding and a reference embedding computed from all tokens preceding sparsification. This reference represents the semantic context active at the pruning point. A consistent pattern emerges: once pruned neurons are re-enabled, the similarity trajectory rapidly realigns with that of the dense model, indicating recovery of the original representational manifold.

Importantly, even without mask release (\textit{Prune@100, No Release}), the similarity curve shows an attempted departure from the initial manifold, demonstrating that the model’s attention dynamics still signal a domain transition. However, releasing neurons too late (e.g., at the 500th token) produces a distinct failure signature: similarity remains confined to a narrow band, suggesting the generation trajectory has collapsed into a restricted semantic basin. Beyond this point, restoring neurons does not recover the intended trajectory.

\begin{figure}[h]
    \centering

    \begin{subfigure}{0.48\textwidth}
        \includegraphics[width=\linewidth]{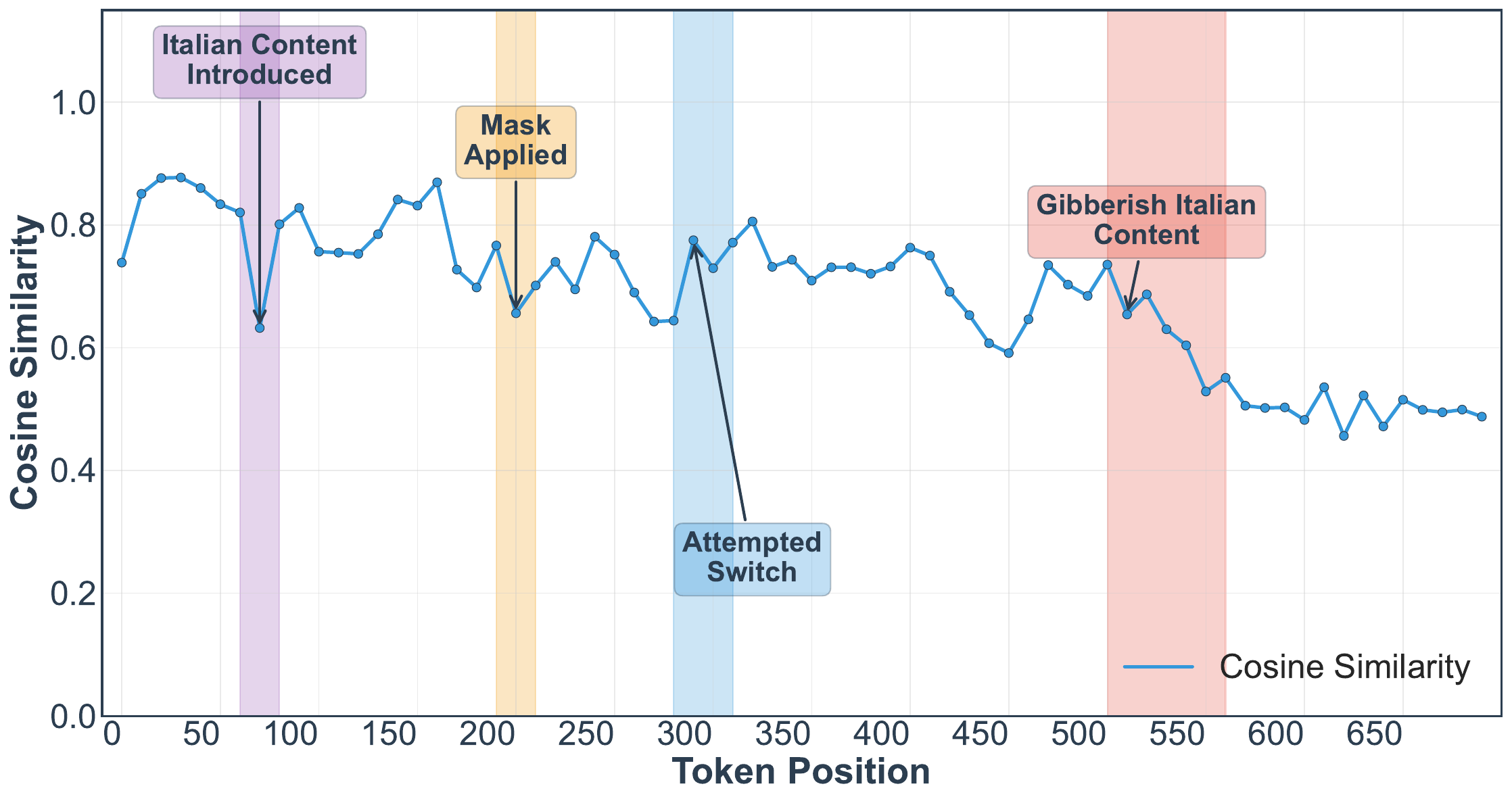}
        \caption{Model pruned at 100th generated token}
    \end{subfigure}\hfill
    \begin{subfigure}{0.48\textwidth}
        \includegraphics[width=\linewidth]{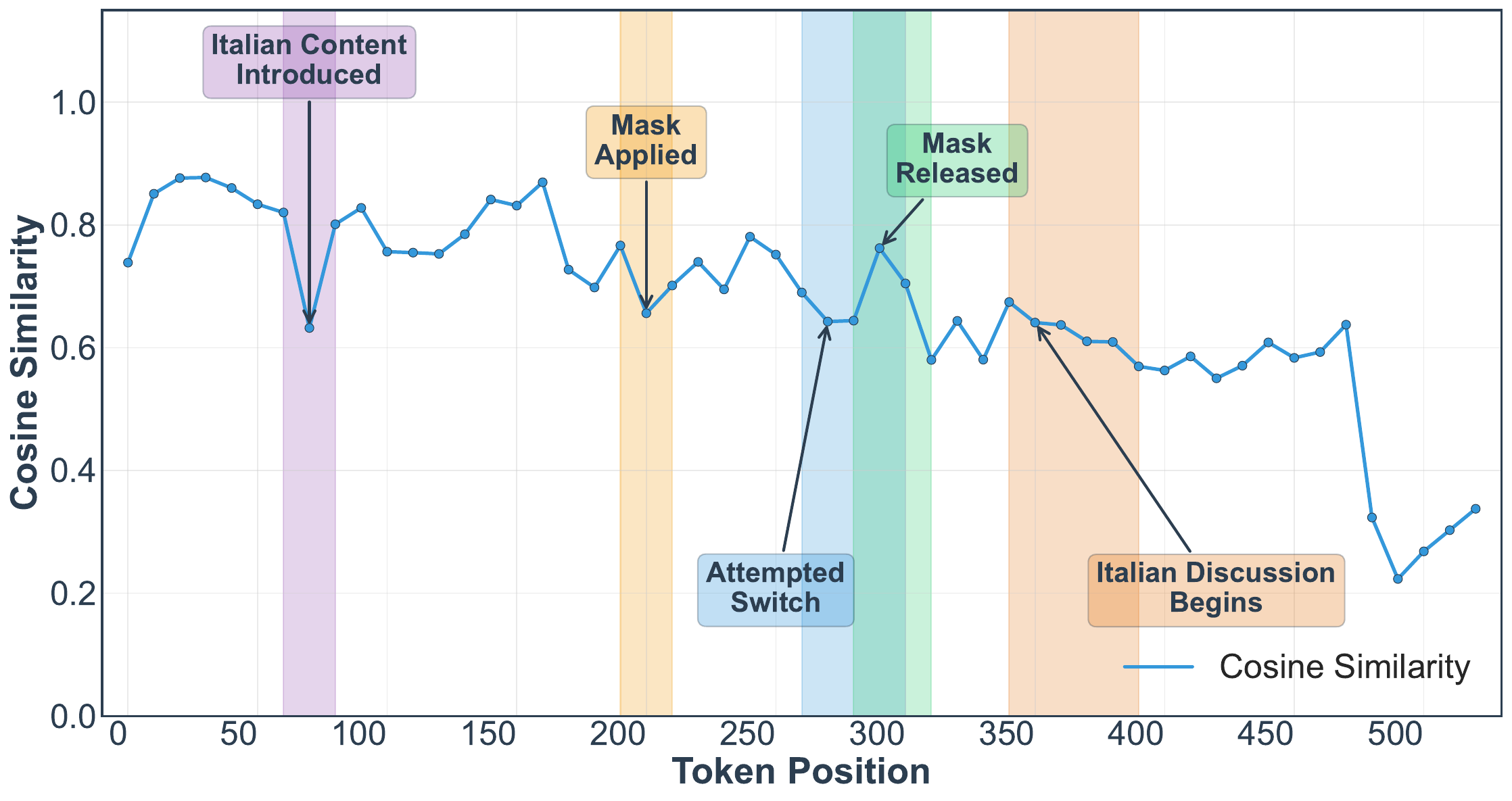}
        \caption{Pruned neurons released after the 200th generated token}
    \end{subfigure}

    \vspace{0.3cm} 

    \begin{subfigure}{0.48\textwidth}
        \includegraphics[width=\linewidth]{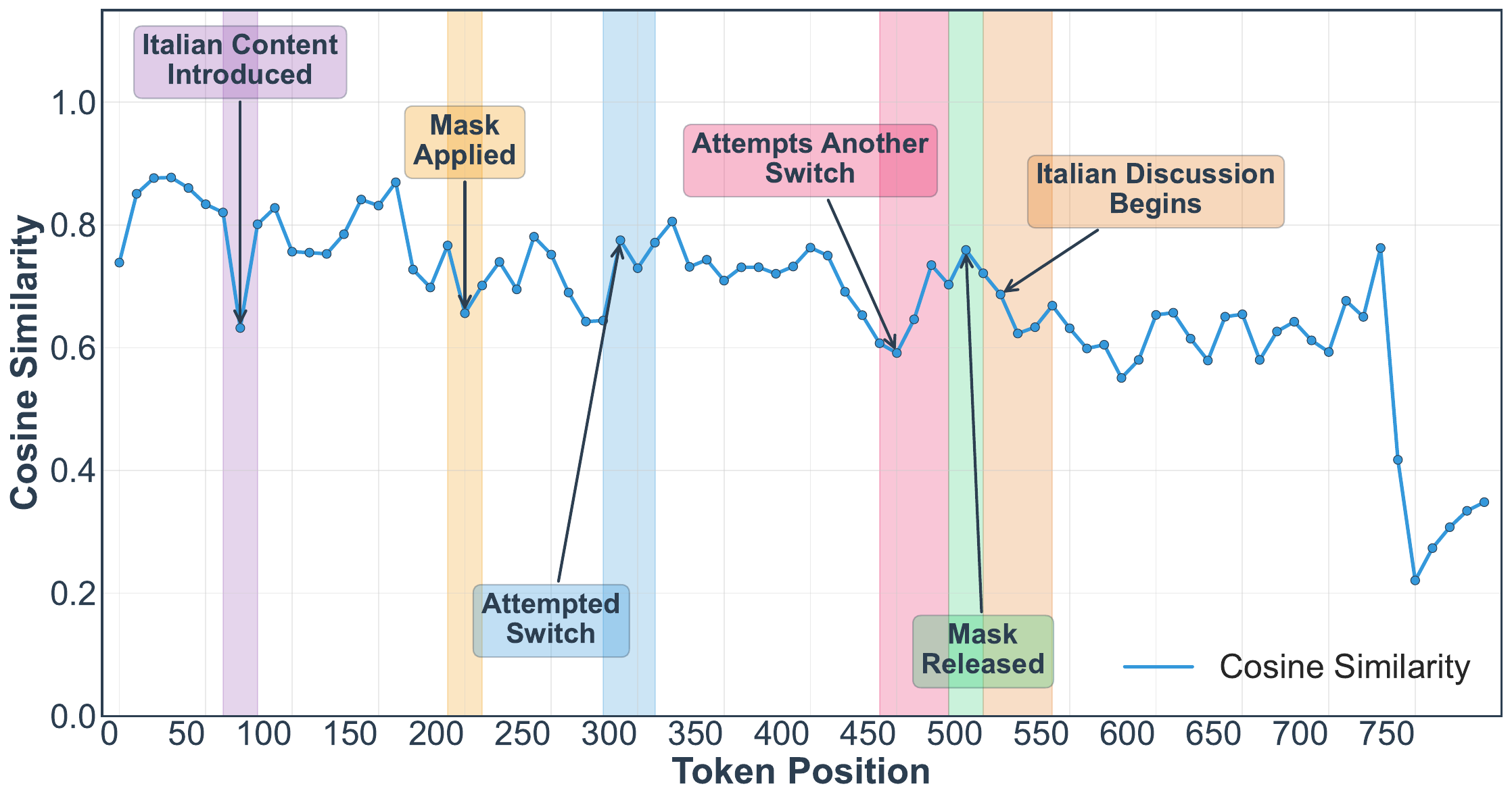}
        \caption{Pruned neurons released after the 400th generated token}
    \end{subfigure}\hfill
    \begin{subfigure}{0.48\textwidth}
        \includegraphics[width=\linewidth]{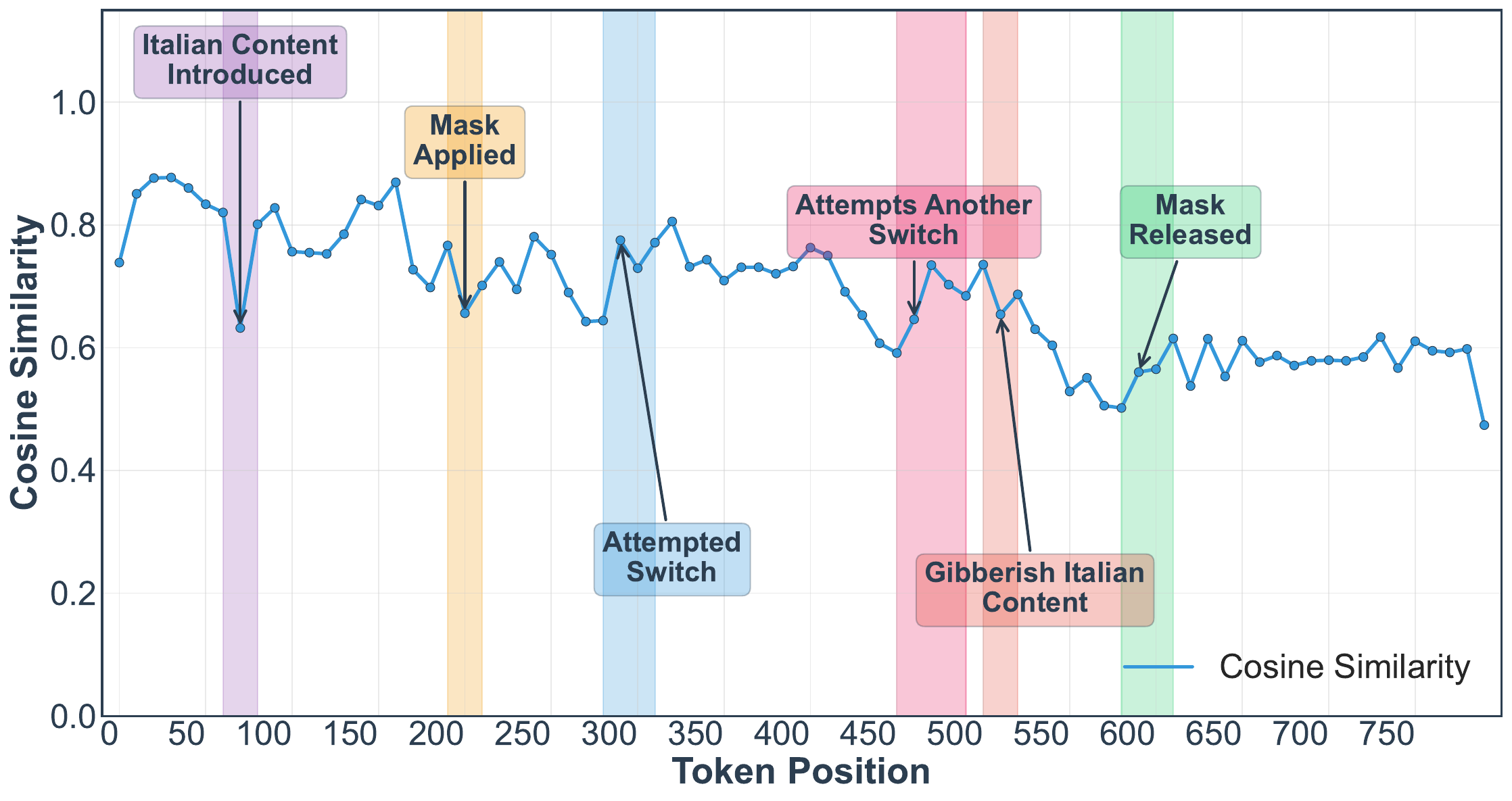}
        \caption{Pruned neurons released after the 500th generated token}
    \end{subfigure}

    \caption{Cosine similarity trajectories under different pruning strategies. Each subplot shows the effect of neuron pruning and release timing on the model's internal representations.}
    \label{fig:cosine-sim-four-plots}
\end{figure}

Fine-grained cases in \cref{fig:cosine-sim-four-plots} further illustrate this mechanism. In panels (b) and (c), timely neuron restoration enables a sustained shift in similarity, corresponding to successful transition to the second topic. When neurons remain unavailable, the trajectory repeatedly reverts toward the original manifold, indicating that the model lacks the representational capacity to sustain the new domain. In panel (d), late intervention fails to alter the trajectory, confirming that knowledge drift has become irreversible.

Overall, attention-space divergence provides an early signal of domain shift attempts, while the persistence of low-diversity similarity trajectories indicates semantic trapping. Successful generation therefore depends on synchronizing neuron availability with the evolving attention-driven knowledge trajectory.

\clearpage
\newpage
\section{Custom prompt dataset}

We did not identify any standard benchmark containing prompts that explicitly require multi-topic generation within a single response, which is necessary to evaluate generation under pruning with knowledge tracing. To address this gap, we designed a set of prompt templates with placeholders that can be instantiated using different topic variables, enabling systematic construction of diverse multi-domain prompts. By sampling across these placeholders, we generated 500 distinct prompts covering varied topic combinations and transitions.

\begin{sidewaysfigure}
    \centering
    \includegraphics[width=\textheight, height=\textwidth, keepaspectratio]{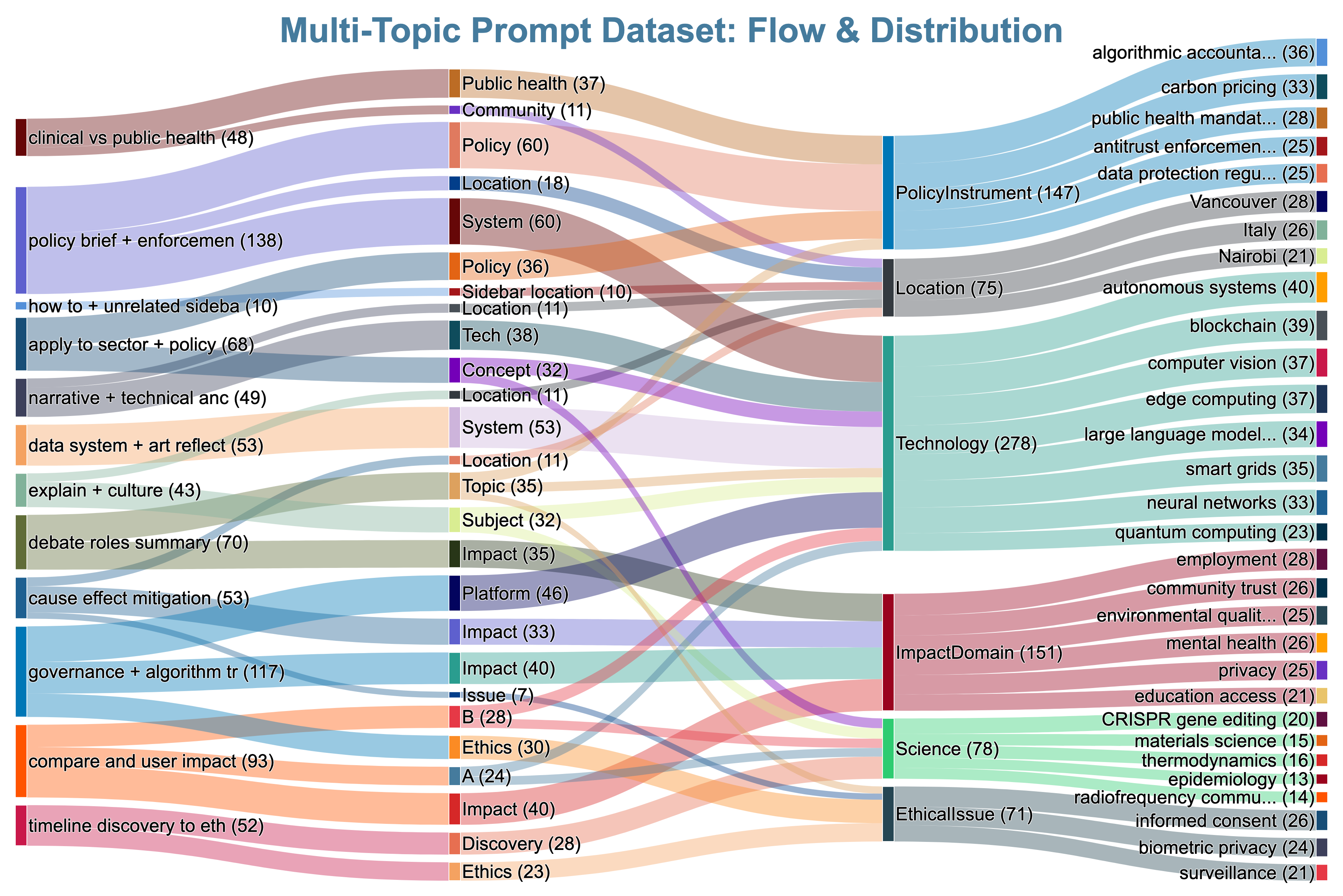}
    \caption{Distribution of generated prompts over the templates and subjects. The number next to each label signifies the number of prompt using the element.}
    \label{prompts_sankey}
\end{sidewaysfigure}

Table~\ref{prompts_table} summarizes the templates, their intended evaluation purpose, and the categories of terms used to populate each placeholder. This design ensures controlled variation in domain shifts while maintaining consistent structural demands on the model. The Sankey diagram in \cref{prompts_sankey} visualizes the distribution of generated prompts across template types and topic categories, confirming broad coverage of cross-domain transitions.

\clearpage
\begingroup
\definecolor{headerblue}{RGB}{0,64,128}
\definecolor{rowgray}{RGB}{242,242,242}
\definecolor{textblue}{RGB}{0,64,128}
\definecolor{slotbg}{RGB}{255,235,153}
\definecolor{slottext}{RGB}{204,102,0}

\newcolumntype{L}[1]{>{\raggedright\arraybackslash}p{#1}}

\newcommand{\slot}[1]{{\setlength{\fboxsep}{1pt}\colorbox{slotbg}{\textcolor{slottext}{\texttt{\{#1\}}}}}}

\setlength{\tabcolsep}{4pt}
\renewcommand{\arraystretch}{1.0}

\rowcolors{2}{rowgray}{white}
\footnotesize
\begin{longtable}{|L{3.2cm}|L{8.2cm}|L{4.5cm}|}
\caption{Intent and Prompt Templates} \label{prompts_table} \\
\hline
\rowcolor{headerblue} 
\textcolor{white}{\textbf{Intent}} & \textcolor{white}{\textbf{Prompt Template}} & \textcolor{white}{\textbf{Slot Values}} \\
\hline
\endfirsthead

\hline
\rowcolor{headerblue}
\textcolor{white}{\textbf{Intent}} & \textcolor{white}{\textbf{Prompt Template}} & \textcolor{white}{\textbf{Slot Values}} \\
\hline
\endhead

\hline
\endfoot

\hline
\endlastfoot

Explain and Contrast & 
\ttfamily Write about \slot{SUBJECT} - what it is, how it works, and typical uses. Then separately write about \slot{CULTURE} in \slot{LOCATION}, covering traditions, variations, and social context. & 
{\textbf{\color{textblue}SUBJECT:} Technology, Science 

\textbf{\color{textblue}CULTURE:} CulturalPractice 

\textbf{\color{textblue}LOCATION:} Region, Country} \\
\hline

Compare and Assess Impact & 
\ttfamily Compare \slot{A} and \slot{B}, focusing on principles, performance, and trade-offs. Then explain how these differences affect \slot{IMPACT} for \slot{GROUP}. & 
{\textbf{\color{textblue}A:} Technology, Science, EconomicConcept 

\textbf{\color{textblue}B:} Technology, Science, EconomicConcept 

\textbf{\color{textblue}IMPACT:} ImpactDomain 

\textbf{\color{textblue}GROUP:} PopulationGroup} \\
\hline

Apply to Sector with Policy & 
\ttfamily Write about how \slot{CONCEPT} applies to \slot{SECTOR}, including workflow changes, benefits, and risks. Then discuss policy measures such as \slot{POLICY} that shape adoption. & 
{\textbf{\color{textblue}CONCEPT:} Technology, Science 

\textbf{\color{textblue}SECTOR:} Sector 

\textbf{\color{textblue}POLICY:} PolicyInstrument} \\
\hline

Analyze Cause, Effect, and Mitigation & 
\ttfamily Describe how \slot{ISSUE} affects \slot{IMPACT} in \slot{LOCATION}, covering mechanisms and evidence. Then propose practical mitigation strategies and trade-offs. & 
{\textbf{\color{textblue}ISSUE:} EnvironmentIssue, HealthCondition, EthicalIssue 

\textbf{\color{textblue}IMPACT:} ImpactDomain 

\textbf{\color{textblue}LOCATION:} Country, City, Region, RuralArea} \\
\hline

Debate with Role-Play & 
\ttfamily Create a dialogue between a \slot{ROLE\_A} and a \slot{ROLE\_B} discussing \slot{TOPIC}. Start with the \slot{ROLE\_A}'s perspective. End with a summary addressing \slot{IMPACT}. & 
{\textbf{\color{textblue}ROLE\_A:} ActorRole 

\textbf{\color{textblue}ROLE\_B:} ActorRole 

\textbf{\color{textblue}TOPIC:} EthicalIssue, Technology, PolicyInstrument 

\textbf{\color{textblue}IMPACT:} ImpactDomain} \\
\hline

Trace Development and Ethics & 
\ttfamily Trace the scientific development of \slot{DISCOVERY}. Then explain its commercialization in \slot{SECTOR}. Finally, analyze ethical concerns focusing on \slot{ETHICS}. & 
{\textbf{\color{textblue}DISCOVERY:} Science 

\textbf{\color{textblue}SECTOR:} Sector 

\textbf{\color{textblue}ETHICS:} EthicalIssue} \\
\hline

Narrative with Technical Explanation & 
\ttfamily Write a short narrative about \slot{EVENT\_GROUP} in \slot{LOCATION}. Then add a factual explanation of how \slot{TECH} enabled or shaped this experience. & 
{\textbf{\color{textblue}EVENT\_GROUP:} PopulationGroup 

\textbf{\color{textblue}TECH:} Technology 

\textbf{\color{textblue}LOCATION:} Country, City, Region, RuralArea} \\
\hline

Policy Brief with Enforcement & 
\ttfamily Draft a policy brief addressing \slot{POLICY} for \slot{SYSTEM} in \slot{LOCATION}. Then detail technical enforcement and auditability mechanisms. & 
{\textbf{\color{textblue}POLICY:} PolicyInstrument 

\textbf{\color{textblue}SYSTEM:} Technology 

\textbf{\color{textblue}LOCATION:} Country, City} \\
\hline

Governance and Transparency & 
\ttfamily Outline governance principles for \slot{PLATFORM}, accounting for \slot{ETHICS}. Then explain how transparency in ranking and recommendation algorithms supports \slot{IMPACT}. & 
{\textbf{\color{textblue}PLATFORM:} Technology 

\textbf{\color{textblue}ETHICS:} EthicalIssue 

\textbf{\color{textblue}IMPACT:} ImpactDomain} \\
\hline

How-To Guide with Sidebar & 
\ttfamily Provide a concise how-to guide for \slot{WORKFLOW}. Then add a sidebar describing \slot{SIDEBAR\_CULTURE} in \slot{SIDEBAR\_LOCATION}. & 
{\textbf{\color{textblue}WORKFLOW:} Sector, EducationTheme 

\textbf{\color{textblue}SIDEBAR\_LOCATION:} Country, Region, City, RuralArea 

\textbf{\color{textblue}SIDEBAR\_CULTURE:} CulturalPractice} \\
\hline

System Explanation with Artistic Reflection & 
\ttfamily Explain how \slot{SYSTEM} processes inputs and produces outputs. Then reflect on parallels with \slot{ART}, considering structure, style, and interpretation. & 
{\textbf{\color{textblue}SYSTEM:} Technology 

\textbf{\color{textblue}ART:} ArtForm} \\
\hline

Compare Clinical and Public Health Approaches & 
\ttfamily Contrast clinical approaches to \slot{CLINICAL} with population-level interventions like \slot{PUBLIC\_HEALTH}. Then discuss how local norms in \slot{COMMUNITY} shape outcomes. & 
{\textbf{\color{textblue}CLINICAL:} HealthCondition 

\textbf{\color{textblue}PUBLIC\_HEALTH:} PolicyInstrument 

\textbf{\color{textblue}COMMUNITY:} City, Region, Country} \\
\hline

\end{longtable}
\endgroup

\clearpage
\newpage
\section{System-Level Impact of FFN Pruning}
Wall-clock acceleration depends heavily on hardware-specific factors such as kernel fusion, memory hierarchy behavior, runtime scheduling, and vendor-optimized libraries, making such numbers difficult to generalize across systems. Instead, we report the exact memory traffic and compute requirements, which are hardware-agnostic quantities directly reflecting algorithmic demand. These metrics more accurately characterize how sparsification reshapes system bottlenecks and enable fair comparison across platforms, accelerators, and architectural simulators.

Table~\ref{tab:compute_mem_compare} highlights a key but often overlooked property of structured FFN sparsification: it does not uniformly shrink the transformer, but instead selectively removes the dominant system bottleneck. For \textsc{Llama-3.1-70B}, dense execution shows the MLP consuming 53.91\,GB of memory traffic per layer versus only 15.08\,GB for attention, making the FFN responsible for nearly 78\% of total layer-wise memory movement. After 70\% FFN sparsification, this drops to 17.14\,GB, bringing MLP and attention to comparable bandwidth demands. A similar trend holds for compute: MLP FLOPs reduce from 13.12 to 4.12 TFLOPs, rebalancing the layer’s arithmetic distribution relative to attention (2.85 TFLOPs). This shift transforms the execution regime from MLP-dominated to a more balanced profile, which is critical because decode inference is typically memory-bandwidth-bound rather than compute-bound. 

From a systems perspective, FFN pruning therefore acts as a bottleneck rebalancing mechanism. It reduces the dominant source of memory traffic, lowers per-token energy and bandwidth requirements, and smooths inter-layer workload distribution without altering attention’s sequence-dependent behavior. This property is particularly significant for large-batch decode workloads and distributed or NoC-based accelerators, where memory movement — not parameter count alone — dictates scalability and efficiency.

\begin{table*}[h]
\centering
\caption{Attention and MLP cost comparison across model scales under dense and 70\% sparse configurations at FP8 compute precision and FP16 communication precision.}
\label{tab:compute_mem_compare}
\begin{sc}{%
\begin{tabular}{ll|cc|cc}
\toprule
\multirow{2}{*}{Component} & \multirow{2}{*}{Metric}
& \multicolumn{2}{c|}{\textsc{LLaMA-3.1-70b}} 
& \multicolumn{2}{c}{\textsc{LLaMA-3.1-8b}} \\
\cmidrule(lr){3-4} \cmidrule(lr){5-6}
& & Dense & Sparse (70\%) & Dense & Sparse (70\%) \\
\midrule
\multirow{2}{*}{Attention}
& Memory Access (GB) & 15.08 & 15.08 & 2.53 & 2.53 \\
& Compute (TFLOPs)   & 2.85  & 2.85  & 0.32 & 0.32 \\
\midrule
\multirow{2}{*}{MLP}
& Memory Access (GB) & 53.91 & 17.14 & 5.53 & 1.50 \\
& Compute (TFLOPs)   & 13.12 & 4.12  & 1.31 & 0.35 \\
\bottomrule
\end{tabular}%
}\end{sc}
\end{table*}

\end{document}